\documentclass[manuscript,screen,nonacm]{acmart}

\AtBeginDocument{%
  }

\usepackage{multirow}
\usepackage{subfigure}
\usepackage{array}

\begin{document}
\renewcommand\footnotetextcopyrightpermission[1]{} 



\title[A Survey of What to Share in Federated Learning]{A Survey of What to Share in Federated Learning: Perspectives on Model Utility, Privacy Leakage, and Communication Efficiency}


\author{Jiawei Shao}
\authornote{Authors contributed equally to this research.}
\email{jiawei.shao@connect.ust.hk}
\orcid{0000-0001-8836-1430}

\author{Zijian Li}
\authornotemark[1]
\email{zijian.li@connect.ust.hk}
\orcid{0000-0001-5094-3697}

\author{Wenqiang Sun}
\email{wsunap@connect.ust.hk}
\orcid{0009-0008-7990-7831}
\authornotemark[1]

\author{Tailin Zhou}
\email{tzhouaq@connect.ust.hk}
\orcid{0000-0002-1167-9392}

\author{Yuchang Sun}
\email{yuchang.sun@connect.ust.hk}
\orcid{0000-0001-7881-4723}

\author{Lumin Liu}
\email{lliubb@connect.ust.hk}
\orcid{0000-0001-8878-8901}

\author{Zehong Lin}
\email{eezhlin@ust.hk}
\orcid{0000-0002-9503-2464}

\affiliation{%
  \institution{Hong Kong University of Science and Technology}
  \city{Hong Kong}
  \country{China}
}

\author{Yuyi Mao}
\email{yuyi-eie.mao@polyu.edu.hk}
\orcid{0000-0002-5646-8679}
\affiliation{%
  \institution{Hong Kong Polytechnic University}
  \city{Hong Kong}
  \country{China}
}

\author{Jun Zhang}
\email{eejzhang@ust.hk}
\orcid{0000-0002-5222-1898}
\affiliation{%
  \institution{Hong Kong University of Science and Technology}
  \city{Hong Kong}
  \country{China}
}

\authorsaddresses{Authors' addresses: J. Shao, Z. Li, W. Sun, T. Zhou, Y. Sun, L. Liu, Z. Lin, and J. Zhang are with the Department of Electronic and Computer Engineering, Hong Kong University of Science and Technology, China; emails: \{jiawei.shao, zijian.li, wsunap, tzhouaq, yuchang.sun, lliubb\}@connect.ust.hk, \{eezhlin, eejzhang\}@ust.hk; Y. Mao is with the Department of Electrical and Electronic Engineering, Hong Kong Polytechnic University, China; email: yuyi-eie.mao@polyu.edu.hk; The corresponding author is J. Zhang.}


\renewcommand{\shortauthors}{J. Shao, et al.}

\begin{abstract}
Federated learning (FL) has emerged as a secure paradigm for collaborative training among clients. 
Without data centralization, FL allows clients to share local information in a privacy-preserving manner.
This approach has gained considerable attention, promoting numerous surveys to summarize the related works.
However, the majority of these surveys concentrate on FL methods that share model parameters during the training process, while overlooking the possibility of sharing local information in other forms.
In this paper, we present a systematic survey from a new perspective of \emph{what to share} in FL, with an emphasis on the model utility, privacy leakage, and communication efficiency.
First, we present a new taxonomy of FL methods in terms of three sharing methods, which respectively share model, synthetic data, and knowledge.
Second, we analyze the vulnerability of different sharing methods to privacy attacks and review the defense mechanisms.
Third, we conduct extensive experiments to compare the learning performance and communication overhead of various sharing methods in FL.
Besides, we assess the potential privacy leakage through model inversion and membership inference attacks, while comparing the effectiveness of various defense approaches.
Finally, we identify future research directions and conclude the survey.
\end{abstract}








\maketitle

\section{Introduction}

Artificial intelligence (AI) has been successfully integrated into a wide variety of applications, such as natural language processing \cite{chowdhary2020natural}, computer vision \cite{voulodimos2018deep}, and speech recognition \cite{zhang2018deep}, resulting in remarkable performance enhancements.
The success of AI systems depends on two key ingredients: massive datasets and powerful computing capabilities.
Conventionally, centralized training is used for building high-performing models, where all data are consolidated in a central server. 
However, this approach strongly relies on data collection, which poses several challenges, including data availability and data privacy.
On one hand, due to the competitive industry landscape, data generally exist in the form of isolated islands, which makes it expensive and time-consuming to break down the barriers between data sources.
On the other hand, with the growing emphasis on data security and user privacy, regulations such as the General Data Protection Regulation (GDPR) \cite{voigt2017eu_GDPR} and California Consumer Privacy Act (CCPA) \cite{pardau2018california_CCPA} have been enforced to restrict data collection and storage.

\begin{figure*}[t]
\centering

\subfigure[Model sharing]{
\centering
\includegraphics[width=0.31\linewidth]{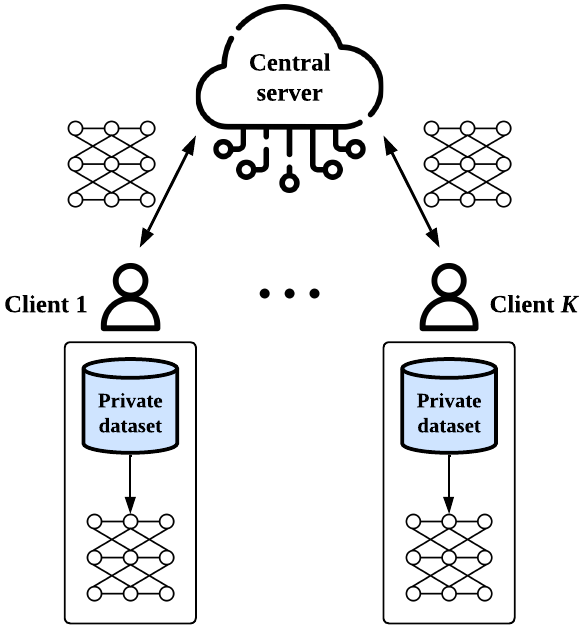}
}
\subfigure[Synthetic data sharing]{
\centering
\includegraphics[width=0.31\linewidth]{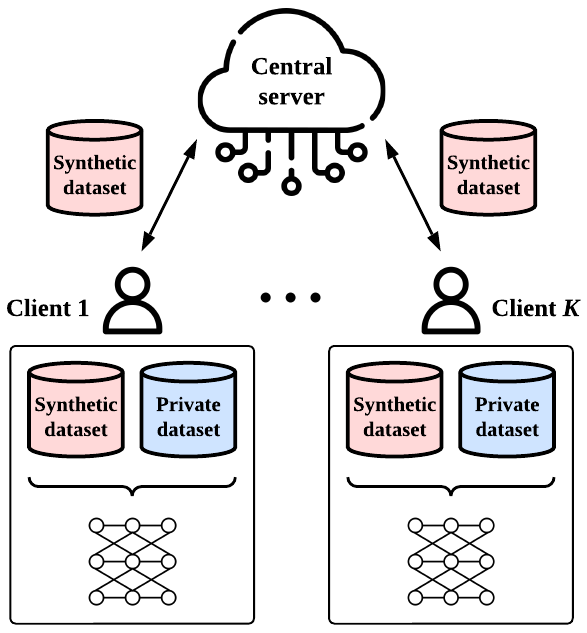}
}
\subfigure[Knowledge sharing]{
\centering
\includegraphics[width=0.31\linewidth]{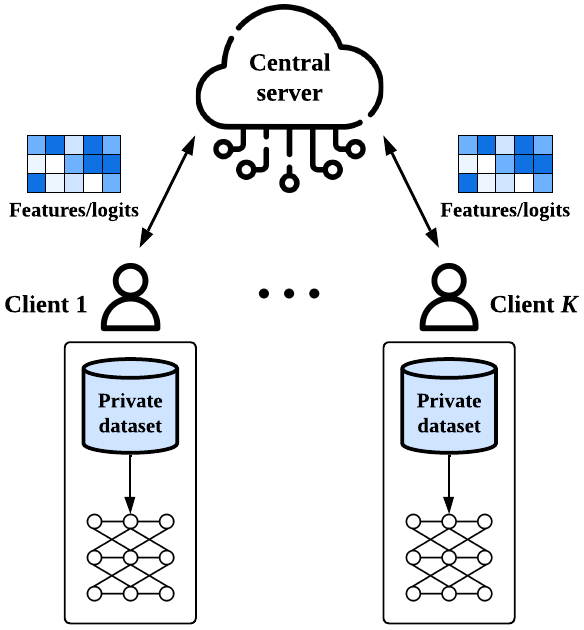}
}
\caption{An overview of different sharing methods in FL.}
\label{Fig:sharing_methods_FL}
\end{figure*}

To address the above challenges, federated learning (FL) \cite{fedavg} has emerged as a promising solution by enabling distributed model training without requiring the sharing of raw data.
This is achieved by deploying a central server to coordinate multiple clients that iteratively optimize their local models on their local datasets and share \emph{privacy-preserving information}.
The compilation of the shared information represents the collective expertise gained from the private datasets, which is a key determinant of the federated training performance.
The shared information in the literature of FL can be categorized into three main types, namely, model parameters, synthetic data, and knowledge, as depicted in Fig. \ref{Fig:sharing_methods_FL}, leading to the following three sharing methods:

\begin{itemize}
\item \textbf{Model sharing:}
Clients share local model parameters or updates to the server for aggregation.
The aggregated model is sent back to the clients for further optimization based on local data.

\item \textbf{Synthetic data sharing:} Clients generate and share synthetic data that are dissimilar to raw data but retain important features to train local models.

\item \textbf{Knowledge sharing:} Clients share layer activations and use them to update local models via knowledge distillation (KD) \cite{gou2021knowledge}.

\end{itemize}

These sharing methods enable FL to leverage the collective intelligence of distributed clients without exposing individual data points, making it a promising approach for large-scale collaborative machine learning.
The efficacy of FL has been validated in various domains \cite{huang2023federated, xu2021federated, long2020federated, jiang2020federated}, such as healthcare, financial services, and smart cities. 
Nevertheless, despite its potential, FL still faces three primary challenges:
\begin{itemize}
\item \textbf{Data heterogeneity:} The basic assumption of machine learning is that the training samples are independent and identically distributed (IID).
This assumption, however, does not hold in the context of FL, as clients collect data in different environments and contexts, leading to non-IID distributions.
This data heterogeneity can result in poor convergence or a slower learning process \cite{fedprox}.
\item \textbf{Privacy leakage:} While FL is intended to enhance privacy protection, recent studies \cite{geiping2020inverting,fredrikson2015model} have demonstrated that it is still vulnerable to privacy attacks.
For instance, model inversion attacks \cite{fredrikson2015model} can reconstruct private training data from the shared model.
\item \textbf{Communication overhead:} 
FL involves the frequent exchange of local information between clients and a server, which results in considerable communication overhead \cite{sun2021semi,sun2023semi,sun2022asynchronous}.
Such an issue is particularly challenging when collaboratively training large-scale models \cite{bubeck2023sparks, kasneci2023chatgpt} by sharing model updates.

\end{itemize}

As discussed in \cite{zhang2022trading_tradeoff_QYang}, there exists a tradeoff between utility (i.e., model performance), privacy, and efficiency in FL.
The sharing methods critically affect these three factors.
For example, sharing knowledge can significantly reduce communication overhead and protect the local models against white-box attacks.
Nonetheless, this method may not achieve the desired model performance, especially when the data are highly non-IID.
Moreover, when sharing synthetic samples, there is an inherent tradeoff between the model utility and the potential breach of privacy. 
If the synthetic samples preserve more information from the original samples, there can be a more significant improvement in model performance. 
This, however, also increases the risk of privacy breaches.
Recently, many efforts \cite{shen2022share,kim2021federated} have been made to improve the training performance while providing high privacy guarantees and reducing communication costs.
In this survey, we will take a closer look at these endeavors and provide a systematic analysis of \emph{what to share} in FL.

\subsection{Related Surveys and Our Contributions}

The pioneering work \cite{yang2019federated} introduced the basic concepts of FL.
Subsequently, Li et al. \cite{li2020federated} investigated the unique properties of FL compared with distributed computation and privacy-preserving learning approaches. 
In addition, the study \cite{li2021survey} presented an overview of the FL systems and pointed out the design factors for successful federated training.

Despite the rising popularity of FL, Kairouz et al. \cite{kairouz2021advances} have identified a collection of open problems that need to be addressed, including non-IID data, privacy risks, and limited communication bandwidth.
Several studies have expanded the discussion on these issues and summarized potential solutions \cite{li2022federated,lyu2020threats,rodriguez2023survey,lim2020federated}.
For instance, Li et al. \cite{li2022federated} reviewed FL algorithms tailored
to address the learning effectiveness under non-IID data settings and conducted extensive experiments to evaluate their performance.
The survey conducted by Lyu et al. \cite{lyu2020threats} focused on security and privacy concerns, and Rodriguez et al. \cite{rodriguez2023survey} conducted experiments to evaluate the vulnerability of both adversarial and privacy attacks.
Moreover, \cite{lim2020federated} considered the communication challenges in FL and reviewed potential solutions to reduce communication costs.
Nonetheless, the majority of these surveys concentrated on model sharing, while neglecting the opportunities to improve FL by sharing information in other forms.

Recently, synthetic data and knowledge sharing have attracted increasing attention \cite{zhu2021federated,ma2022state,wu2023survey}. 
Authors of \cite{zhu2021federated} summarized the existing data-based and knowledge-based methods to mitigate the non-IID problem.
Wu et al. \cite{wu2023survey} made the first attempt to investigate the capability of knowledge sharing methods to tackle the challenges caused by system heterogeneity.
However, existing literature reviews lack comprehensive analysis and empirical evidence on how different sharing methods affect model utility, privacy leakage, and communication efficiency in FL.

In this manuscript, we aim to fill these gaps by conducting a thorough survey and analysis of FL methods that involve sharing different types of information.
The comparison of our study with existing survey articles is summarized in Table \ref{table:related_surveys}, and the novel contributions are as follows:

\begin{table*}[]
\centering
\footnotesize
\caption{Summary of Related Surveys}
\setlength\tabcolsep{2pt}
\begin{tabular}{c|ccc|cccc|m{0.55\textwidth}}
\hline
\multirow{2}{*}{Ref.} & \multicolumn{3}{c|}{What to share}            & \multirow{2}{*}{\begin{tabular}[c]{@{}c@{}}Model\\ utility\end{tabular}} & \multirow{2}{*}{\begin{tabular}[c]{@{}c@{}}Privacy\\ leakage\end{tabular}} &  \multirow{2}{*}{\begin{tabular}[c]{@{}c@{}}Commun.\\ efficiency\end{tabular}} & \multirow{2}{*}{\begin{tabular}[c]{@{}c@{}}Exp.\\ results\end{tabular}} &\multirow{2}{*}{\begin{tabular}[c]{@{}c@{}}Contributions\end{tabular}} \\
                      & Model & Data & Knowl. &                                &                                  &                                           &                                    \\ \hline
\cite{yang2019federated}                      & $\checkmark$     &                &           &  $\checkmark$                              &  $\checkmark$                                &                                           &                        &    Introduced a comprehensive secure federated learning framework, including horizontal federated learning, vertical federated learning, and federated transfer learning.
        \\ \hline
\cite{li2020federated}                      &  $\checkmark$   &                &           &   $\checkmark$                            &   $\checkmark$                              &   $\checkmark$                                        &                &    Discussed the types, architecture, and applications of FL. Identified various challenges due to a heterogeneous and distributed environment.
                \\ \hline
\cite{li2021survey}                      & $\checkmark$     &                &           &  $\checkmark$                              &     $\checkmark$                             &  $\checkmark$                                         &                &    Categorized federated learning systems according to six different aspects, including data distribution, machine learning model, privacy mechanism, communication architecture, scale of federation, and motivation of federation.
                    \\ \hline
\cite{kairouz2021advances}                      & $\checkmark$     &                &           &  $\checkmark$                              &     $\checkmark$                             &    $\checkmark$                                       &          &    Presented an extensive collection of open problems and challenges in FL, including training efficiency, privacy threats, adversarial attacks, fairness, and system constraints.
                          \\ \hline
\cite{li2022federated} & $\checkmark$     &                &           & $\checkmark$                                &                                  &                                           &                                   $\checkmark$    &   Presented comprehensive data partitioning strategies to cover the typical non-IID data cases in FL and conducted extensive experiments to evaluate different FL algorithms.\\ \hline
\cite{lyu2020threats}                      & $\checkmark$     &                &           &                                &  $\checkmark$                                &                                           &                                       &     Surveyed the threats to compromise FL, specifically focusing on poisoning attacks and inference attacks.
 \\ \hline
\cite{rodriguez2023survey}                      & $\checkmark$     &                &           &   $\checkmark$                             &   $\checkmark$                               &                                           &    $\checkmark$      &     Provided a taxonomy of adversarial attacks and a taxonomy of defense methods that depict a general picture of the vulnerability of federated learning, and conducted experiments to study the performance of attacks and defenses.
                     \\ \hline
\cite{lim2020federated}                      & $\checkmark$     &                &           &   $\checkmark$                             &   $\checkmark$                               &   $\checkmark$                                        &             &     Introduced how FL can serve as an enabling technology for collaborative model training at mobile edge networks, and discussed the potential challenges, including communication costs, resource allocation, data privacy, and data security.
                   \\ \hline
\cite{zhu2021federated}                      & $\checkmark$     &    $\checkmark$            &    $\checkmark$       &  $\checkmark$                              &                                  &                                           &             &     Analyzed the influence of Non-IID data on both parametric and non-parametric machine learning models in both horizontal and vertical federated learning.
                  \\ \hline
\cite{ma2022state}                      & $\checkmark$     &    $\checkmark$            &           &    $\checkmark$                            &   $\checkmark$                               &  $\checkmark$                                         &      &     Introduced the challenges brought by non-IID data to FL, and analyzed the existing solutions from four dimensions: data, model, algorithm, and framework.
                        \\ \hline
\cite{wu2023survey}                      &                  &                &  $\checkmark$         &   $\checkmark$                             &                                  &     $\checkmark$                                      &      & Studied the strengths and weaknesses of applying KD techniques in FL, focusing on aspects including heterogeneity, efficiency, and performance.
                              \\ \hline
Ours                  & $\checkmark$                 &   $\checkmark$             &  $\checkmark$         &     $\checkmark$                           &    $\checkmark$                              &        $\checkmark$                                   &    $\checkmark$     & Present a survey from a new perspective on what to share in FL, with an emphasis on the model utility, privacy leakage, and communication efficiency. Conduct experiments to empirically evaluate the performance across different sharing methods.
                           \\ \hline
\end{tabular}
\label{table:related_surveys}
\end{table*}

\begin{itemize}

\item We introduce a new taxonomy that classifies FL methods into three categories based on their shared information, namely, model, synthetic data, and knowledge.
For each category, a comprehensive review of the state-of-the-art FL methods is presented.

\item We delve into an in-depth exploration of the potential privacy risks associated with FL and analyze the vulnerability of different sharing methods to privacy attacks. 
A thorough review of defense methods that can mitigate these risks is also provided.

\item We conduct extensive experiments to compare the performance and communication overhead of various sharing methods in FL.
Moreover, we assess the susceptibility of these methods to privacy breaches caused by model inversion and membership inference attacks, and evaluate the effectiveness of different defense methods.

\item We draw valuable insights from our literature review and experimental study. 
These findings shed light on potential research directions and opportunities to improve the FL systems.

\end{itemize}

\subsection{Paper Organization}

\begin{figure*}
    \centering
    \includegraphics[width = 0.95\linewidth]{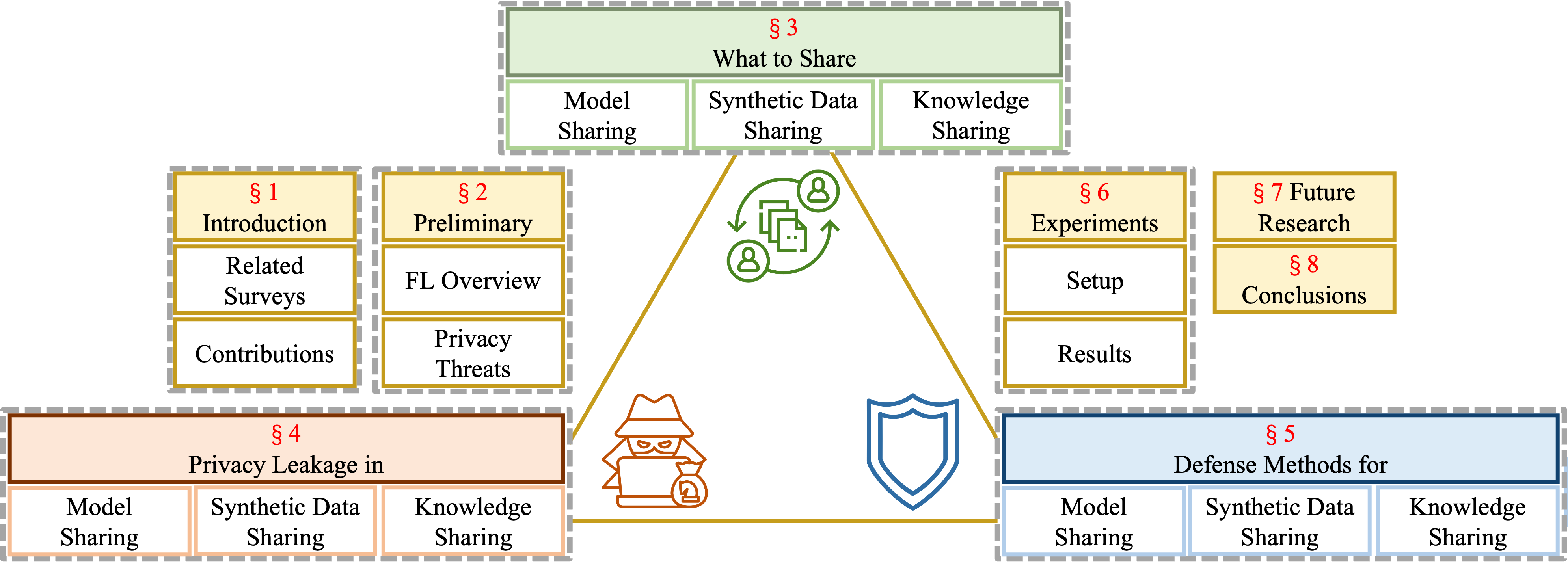}
    \caption{Overview of the survey.}
    \label{fig:overview_survey}
\end{figure*}

The organization of this paper is shown in Fig. \ref{fig:overview_survey}.
Section \ref{sec:preliminary} first introduces the basic concepts of FL and privacy threats.
Section \ref{sec:sharing_method} and Section \ref{sec:privacy_attack} present the taxonomies of sharing methods and privacy attacks in FL, respectively.
Section \ref{sec:defense} reviews the defense methods against privacy attacks.
In Section \ref{sec:exp}, we conduct experimental studies to compare the performance of representative methods.
Finally, Section \ref{sec:discussion} outlines several future directions, and Section \ref{sec:conclusion} concludes this survey.


\section{Preliminary}

\label{sec:preliminary}

To gain a comprehensive understanding of the concepts presented in this survey, it is essential to grasp the fundamental principles of FL as well as its privacy risks, which will be overviewed in this section.

\subsection{An Overview of Federated Learning}

As depicted in Fig. \ref{Fig:sharing_methods_FL}, a typical FL framework consists of a central server and multiple distributed clients (e.g., edge devices or organizations). 
Each client possesses a local model and a private dataset that comprises features and labels. For instance, in an image classification task, the features represent the raw image, while the label denotes the class. 
The local datasets, possibly with personal and sensitive information, cannot be shared with other clients or the server.
Moreover, the local data are collected by the clients and thus are likely non-independent and identically distributed (non-IID) across clients. 
The basic principle of FL is to achieve collaborative training among clients without exposing their local private data.
Specifically, FL aims to train a global model or multiple local models by information sharing, depending on the applications.

In general, the training process of FL can be divided into multiple rounds, and each round contains two phases: the local training phase and the information sharing phase.
In the local training phase, clients train their local models using optimization methods such as stochastic gradient descent (SGD) based on the locally available data.
In the information sharing phase, under the coordination of the central server, clients share certain information to calibrate their local models.
Particularly, the types of shared information differ across FL methods. 
For example, FedAvg \cite{fedavg} requires clients to upload the local models to the central server for aggregation.
In FedMD \cite{li2019fedmd}, clients transmit the predictions of proxy samples as knowledge to the server. 
Additionally, some FL methods designed for computer vision tasks \cite{fedmix,hu2022fedsynth} generate and share synthetic samples among clients.
These synthetic samples are visually dissimilar from raw data to preserve privacy but retain the essential patterns for model training.

\subsection{Privacy Threat Model}

Deep learning models are vulnerable to privacy breaches since they unintentionally memorize specific details about the training data.
The privacy attackers can thus reconstruct the private samples \cite{fredrikson2015model} and infer properties of the training dataset \cite{rezaei2021difficulty} by accessing the model parameters or querying the black-box models.
FL, as a type of deep learning approach, suffers from equal or even higher privacy risks since sharing local information during the training process could inadvertently reveal more sensitive data.

\begin{figure}[t]
\centering
\includegraphics[width=0.95\linewidth]{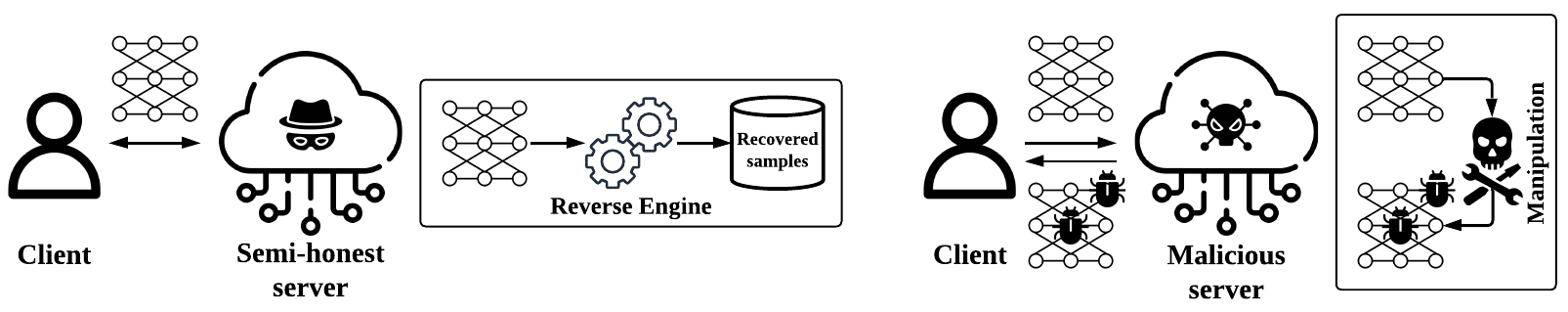}
\caption{Two types of privacy attacks in FL: (Left) passive attack and (Right) active attack.}
\label{Fig:two_types_of_privacy_attacks}
\end{figure}

Fig. \ref{Fig:two_types_of_privacy_attacks} illustrates two typical privacy attacks in FL: passive and active attacks, which are categorized based on the actions taken by the attacker.
In a passive attack, the clients and the server participating in FL are semi-honest.
They follow the FL protocols honestly but attempt to infer private information from the shared information. 
In contrast, in an active attack, the adversaries are malicious participants who intentionally deviate from the FL protocols by altering messages.
This enables them to deceive clients into releasing more private information on local data \cite{hitaj2017deep}.
Fortunately, numerous studies \cite{shejwalkar2021manipulating, fang2020local} have developed defense mechanisms to identify such attackers by monitoring the shared information.
This survey primarily focuses on passive attacks in FL. 
We refer interested readers to the surveys \cite{rodriguez2023survey,lyu2020threats} on active attacks for more comprehensive summaries.

\section{Sharing methods in FL}

\label{sec:sharing_method}

By sharing privacy-preserving information among clients, FL enables collaborative model training on decentralized data while preserving user privacy. 
As the effectiveness and efficiency of FL depend heavily on what to share during the training process, this section classifies FL methods into three categories based on their shared information, namely, model, synthetic data, and knowledge.
A comparison among different sharing methods is shown in Table \ref{table:sharing_methods}.

\begin{table*}[!t]
\centering
\footnotesize
\caption{Comparison among Different Sharing Methods in FL}
\setlength\tabcolsep{2pt}
\begin{tabular}{c|cccccc|m{0.47\textwidth}}
\hline
\multirow{2}{*}{Ref} & \multicolumn{6}{c|}{What to share}                                                              & \multicolumn{1}{c}{\multirow{2}{*}{Description}} \\ 
 \cline{2-7}
                     &     Gradients        &    \begin{tabular}[c]{@{}c@{}}Model\\ parameters\end{tabular}        &   \begin{tabular}[c]{@{}c@{}}Synthetic\\ samples\end{tabular}                 &     \begin{tabular}[c]{@{}c@{}}Generative\\ models\end{tabular}    &      Logits         &    \multicolumn{1}{c|}{Features}          &                              \\ \hline
      
            \cite{chen2016revisiting}         &     \checkmark        &             &                  &                 &               &               &       Iteratively compute local gradients and send them to the server for gradient aggregation.                       \\ \hline
\cite{jhunjhunwala2021adaptive}                     &     \checkmark        &             &                  &                 &               &               &    Quantize local gradient adaptively during the local training and upload the quantized gradient to the server.                          \\ \hline
\cite{shi2019distributed}                     &     \checkmark        &             &                  &                 &               &               &    Compress local gradients based on the top-K sparsification and upload the compressed gradients to the server.                          \\ \hline
\cite{hu2021federated}                     &     \checkmark        &             &                  &                 &               &               &    Combine the random sparsification with gradient perturbation to promote the privacy guarantee and communication efficiency.                    \\ \hline
\cite{fedavg}                     &             &     \checkmark        &                  &                 &               &               &   Train local models via multiple SGD steps and upload model updates for aggregation.                    \\ \hline
\cite{karimireddy2020scaffold}                    &             &     \checkmark        &                  &                 &               &               &    Introduce a regularization term during the local training to mitigate the issue caused by data heterogeneity.                    
\\ \hline
\cite{yurochkin2019bayesian}                   &             &     \checkmark        &                  &                 &               &               &  Propose model aggregation techniques to effectively combine model updates.                    \\ \hline
\cite{data_share_2}                   &             &             &       \checkmark           &                 &               &               &  Share a small portion of local raw data among clients to mitigate the non-IID problem.                    \\ \hline
\cite{mix2fld}                   &             &             &       \checkmark           &                 &               &               &  Generate synthetic samples based on the Mixup technique and share them to improve the model performance.           \\ \hline
\cite{feddc}                  &             &             &       \checkmark           &                 &               &               &  Synthesize a small but informative dataset based on dataset condensation and send them to the server.                    \\ \hline
\cite{feddm}                 &             &             &       \checkmark           &                 &               &               &  Iteratively generate new synthetic samples based on the global model and share them among clients.                    \\ \hline
\cite{fedmix}                 &             &     \checkmark        &       \checkmark           &                 &               &               &  Iteratively share the averaged local data and model parameters to the server.                    \\ \hline
\cite{li2023feature}        &             &      \checkmark       &    \checkmark              &                 &               &               &  Synthesize local data via model inversion and share both synthetic data and model updates.                    \\ \hline
\cite{VHL}        &             &      \checkmark       &    \checkmark              &                 &               &               &  Build and share virtual homogeneous datasets to calibrate features from heterogeneous clients.                    \\ \hline
\cite{fedgan}                 &             &             &                  &     \checkmark            &               &               &  Collaboratively train a global generative model using local raw data and synthetic dataset.                    \\ \hline
\cite{jeong2018communication_FKD1}                 &             &             &                  &                &       \checkmark        &               &  
Send the local average logits per label to the server and utilize the aggregated logits for model distillation.                    \\ \hline
\cite{wang2022knowledge_PLS}     &             &             &                  &                &       \checkmark        &               &  Propose a predicted logits selection method to aggregate local logits for efficient knowledge distillation.                    \\ \hline
\cite{itahara2021distillation}     &             &             &                  &                &       \checkmark        &               &  Upload the local logits on the proxy dataset and perform model distillation based on the aggregated logits sent from the server.                 \\ \hline
\cite{sattler2021cfd}     &             &             &                  &                &       \checkmark        &               &  Develop a quantization mechanism and a delta coding method to compress the logits.                    \\ \hline
\cite{gong2021ensemble}     &             &             &                  &                &  \checkmark            &     \checkmark          &  Share the class-specific attention maps and final predictions to capture the local knowledge.                    \\ \hline
\cite{he2020group}     &             &             &                  &                &      \checkmark        &     \checkmark          &  Upload the intermediate feature maps and logits to train a global model at the server.    
\\ \hline
\cite{zhu2021data_fedgen}     &             &      \checkmark       &                  &                &      \checkmark        &               &  
Share local logits and models for extra data generation to train a global model at the server.             \\ \hline
\cite{li2023fedcir}     &             &      \checkmark       &                  &                &              &          \checkmark     &  Approximate the global feature distribution at the server for better feature alignment during local training.                    \\ \hline
\end{tabular}
\label{table:sharing_methods}
\end{table*}

\subsection{Model Sharing}

Model sharing is widely used in FL \cite{fedavg,fedprox,chen2016revisiting, mills2023faster, shi2019distributed,dai2021distributed}. 
The basic training framework is distributed stochastic gradient descent \cite{chen2016revisiting, mills2023faster, shi2019distributed,dai2021distributed}.
The clients compute gradients based on their local data, and the server averages these gradients to update the global model. 
However, such a framework requires frequent transmission of gradients.
To mitigate this issue, FedAvg and its variants \cite{fedavg,fedprox} allow clients to train their models via multiple SGD steps.
Instead of gradients, the model parameters are uploaded for aggregation, and the global model is then sent back to the clients for further training.
While this line of work reduces the communication frequency, transmitting models between clients and the server still results in significant communication overhead, particularly when dealing with large-scale models.
Moreover, the heterogeneous data pose additional challenges, leading to slow convergence and long training time.

To speed up the convergence of FL, many researchers \cite{fedprox,karimireddy2020scaffold,feddyn,nguyen2020fast,gao2022feddc,li2021model} focused on tackling model divergence issues caused by data heterogeneity.
For instance, FedProx \cite{fedprox} adopts an $\ell_{2}$ regularization term to penalize local models that are far away from the global model.
SCAFFOLD \cite{karimireddy2020scaffold} introduces a control variate to correct drifts in local training.
FedDyn \cite{feddyn} employs a novel dynamic regularization method for FL, where the local loss is dynamically updated to ensure the asymptotic consistency of client optima with stationary points of the global empirical loss.
Another line of studies \cite{entezarirole, yurochkin2019bayesian, wang2020federated, zhou2023understanding} improved the performance of the global model via aggregation optimization.
The standard aggregation method \cite{fedavg} averages the local models based on weight coordinates.
However, the weight permutation invariance principle \cite{entezarirole,li2022federated} points out that this coordinate-based aggregation is suboptimal, as the neurons in some layers may be mismatched.
To address this issue, Yurochkin et al. \cite{yurochkin2019bayesian} developed probabilistic federated neural matching (PFNM) based on the posterior of a Beta-Bernoulli process (BBP) \cite{thibaux2007hierarchical} to match the weights among local models.
Nonetheless, PFNM is only effective on simple architectures like fully-connected layers.
To achieve the weight alignment goal, Wang et al. \cite{wang2020federated} proposed FedMA, which searches for the best permutation matrices using the Hungarian matching algorithm \cite{kuhn1955hungarian}.

Besides improving training efficiency, model compression methods \cite{wang2023performance,yang2022model} are effective in reducing the size of model updates, thereby minimizing communication costs.
For instance, the authors of \cite{alistarh2017qsgd, reisizadeh2020fedpaq} employed a stochastic quantization method, namely QSGD, to quantize local updates for more efficient transmission.
Jhunjhunwala et al. \cite{jhunjhunwala2021adaptive} proposed an adaptive quantization algorithm called AdaQuantFL to dynamically adjust the quantization level during the training process, effectively reducing both communication costs and quantization distortion.
Apart from quantization, sparsification methods have also been widely used for model compression in FL \cite{shi2019distributed,han2020adaptive,hu2021federated}. 
These methods transform a full gradient into a sparse one by selecting a subset of important elements and setting other insignificant coordinates to zero. 
One popular sparsification technique is the Top-K sparsification \cite{shi2019distributed}, which removes a significant portion of gradients without affecting model convergence. 
In addition, Han et al. \cite{han2020adaptive} proposed an adaptive sparsification method, which determines the degree of sparsity automatically based on the model performance.
Furthermore, Hu et al. \cite{hu2021federated} integrated random sparsification with gradient perturbation on each client to amplify the privacy guarantee while improving communication efficiency.

\subsection{Synthetic Data Sharing}

Besides model sharing methods, sharing data is another intuitive yet effective approach to tackle the heterogeneity among clients and improve training efficiency.
In this research line, Yoshida et al. \cite{data_share_2} proposed a hybrid learning framework that assumes a small portion of clients is willing to share their raw data with others, while Tian et al. 
\cite{tian2021towards} developed a pre-shared data training strategy to collect local data from a random set of clients.
Empirical results from these studies demonstrated that sharing limited raw data can significantly improve overall performance.
However, it is important to note that directly sharing raw data in FL poses a serious threat to privacy.

To mitigate the privacy concerns, sharing the synthetic samples has been suggested as an alternative \cite{mix2fld,xormixup}.
For instance, Mix2FLD \cite{mix2fld}, which builds upon the  Mixup technique \cite{zhangmixup}, allows clients to share \emph{mixup} samples with the server and adopts an inverse engineer to generate inversely-mixup samples for federated distillation. 
This approach involves generating synthetic samples by linearly combining multiple raw samples into one sample, which effectively masks the details of each raw sample.
Similarly, Shin et al. \cite{xormixup} developed an XOR-based mixup data augmentation technique.
In addition, various learning-based methods have been developed for data synthesis in FL, such as dataset distillation \cite{dataset_distillation} and dataset condensation \cite{dataset_condensation}. 
The key idea behind these methods is to synthesize a small yet informative set of samples by learning from the large raw dataset, allowing models trained on the synthetic data to achieve comparable performance to those trained on the raw data. 
By generating a small synthetic dataset locally with these methods, each client in FL can share it with the server in one shot \cite{zhou2020distilled, fedd3, feddc} to significantly reduce communication costs. 
This approach can be further extended to iteratively optimize the local synthetic datasets using the updated global model and share them with the server \cite{feddm, FedMK}.
The iterative optimization process allows the synthetic datasets to be continuously refined, resulting in improved model performance.

Recently, many works combined the model sharing and data sharing methods \cite{fedmix,feddpgan,fedgan,VHL} to further enhance the FL performance.
For instance, the authors of \cite{fedmix} proposed a mean augmented federated learning framework that shares model updates and averaged local data to approximate the loss function of global mixup.
Moreover, FedDPGAN \cite{feddpgan} and FedGAN \cite{fedgan} enable clients to collaboratively train a generative adversarial model (GAN) in a federated framework, where the local datasets can be supplemented with synthetic data to improve model training.
SDA-FL \cite{sda_fl}, on the other hand, allows each client to individually train a generator, which provides synthetic data for all the clients and mitigates the data heterogeneity among clients during the subsequent FL training.
Instead of training a generator, HFMDS-FL \cite{li2023feature} optimizes the synthetic data in model inversion technologies, which reduces the training burden of data synthesis and better resolves the non-IID issue.
Alternatively, VHL \cite{VHL} conducts FL with a virtual homogeneous dataset to calibrate the features from the heterogeneous clients, which can be generated from pure noise shared across clients without any private information.

\subsection{Knowledge Sharing}

In addition to model sharing and synthetic data sharing methods, knowledge distillation (KD) is an effective technique that enables the transfer of knowledge among models.
Federated distillation (FD)~\cite{mora2022knowledge} is an extension of KD tailored to the FL framework.
Typical knowledge exchanged among clients includes logits \cite{jeong2018communication_FKD1,seo202216_FKD2,li2019fedmd,itahara2021distillation} and intermediate features \cite{gong2021ensemble, he2020group}. 
The studies in \cite{jeong2018communication_FKD1,seo202216_FKD2} proposed the FKD method that leverages logits for knowledge sharing.
In particular, the clients periodically transmit the \emph{local average logits} per ground-truth label to the server.
The server then aggregates these logits to produce global logits, which are sent back to the clients to calibrate the local training.
Later, Wang et al. proposed a predicted logits selection (PLS) algorithm~\cite{wang2022knowledge_PLS}.
This method characterizes the relationship between the predictive logits scheduling method and the system performance, and selects significant logits for knowledge sharing.
Such a data-free KD framework offers remarkable communication efficiency compared with traditional model sharing methods.
However, these methods suffer from significant performance drops when the data distributions across clients are highly heterogeneous \cite{shao2023selective}.

To address this issue, many studies \cite{li2019fedmd,itahara2021distillation,chang2019cronus,diakonikolas2017being,sattler2021cfd} utilize a proxy dataset to facilitate the knowledge transfer among clients.
Specifically, clients compute the local logits on the proxy samples, and the server aggregates them to produce the global logits for local distillation.
For example, FedMD~\cite{li2019fedmd} adopts a labeled proxy dataset for model training and knowledge distillation.
The authors of~\cite{itahara2021distillation} proposed an entropy reduction aggregation (ERA) scheme to modify the aggregation step.
The empirical evidence showed that using a small temperature coefficient when applying softmax to the aggregated logits reduces the entropy of global logits and improves convergence.
Moreover, the Cronus method~\cite{chang2019cronus} aggregates the logits at the server following the robust mean estimation algorithm~\cite{diakonikolas2017being} to enhance robustness.
To further reduce the communication overhead, compressed federated distillation (CFD)~\cite{sattler2021cfd} develops a quantization mechanism and a delta coding method to compress the logits.

Besides sharing logits, some methods select intermediate features for knowledge sharing.
For instance, the FedAD~\cite{gong2021ensemble} method employs class-specific attention maps and final predictions to capture the knowledge of local models. The diversity of attention maps across local clients is then exploited to calibrate local models for seeking a consensus among them.
Moreover, FedGKT~\cite{he2020group} reformulates FL as a group knowledge transfer training algorithm.
Specifically, the server adopts the intermediate feature maps, along with the corresponding logits, to train a global model.
The discrepancy between the local and global logits is incorporated as a regularization term within the local loss function to facilitate consistency.

Additionally, some methods combine knowledge sharing and model sharing to enhance FL performance.
One such approach is FedFTG~\cite{zhang2022fine}, which leverages KD to refine the global model on the server side. In this method, the server trains a generator model by incorporating information from the local label distribution and the local models.
Another approach is FedGen~\cite{zhu2021data_fedgen}, which learns a lightweight server-side generator, based on the local label counters and parameters, to ensemble client information in a data-free manner.
This generator is then broadcast to clients and regulates local training by providing learned knowledge as an inductive bias.
In addition, the lightweight server-side generator can be further utilized to estimate the global feature distribution to achieve feature alignment among clients \cite{li2023fedcir}.

\subsection{Summary of Sharing Methods}

This section describes three types of sharing methods in FL in terms of what to share between the server and clients, including model sharing, synthetic data sharing, and knowledge sharing.
Model sharing methods are most common in FL, and many optimization strategies have been proposed to improve training efficiency.
However, such sharing methods still face challenges like high communication overhead and difficulties in generalizing to heterogeneous clients.
To tackle these challenges, synthetic data sharing and knowledge sharing are alternative methods.
Synthetic data sharing involves generating informative but non-real samples from the private dataset.
Sharing these samples among clients mitigates the non-IID problem, reduces the need for frequent communication, and allows clients to initialize personalized models based on their available resources.
However, data generation presents its own issues, such as high computational complexity and the tradeoff between data utility and privacy concerns.
Compared with sharing models and synthetic data, sharing knowledge is easier to implement and has the potential to further reduce the communication overhead.
This is because knowledge generation can be model-agnostic, and the resulting logits or features typically have a much smaller size than the models and synthetic data.
However, as the shared knowledge may not contain sufficient information about private data or models, knowledge sharing methods may experience performance degradation, particularly in the face of data heterogeneity.


\section{Privacy Leakage}

\label{sec:privacy_attack}

While FL enables distributed learning by keeping private data locally, the shared information exposes a broad attack surface. 
In this section, we will provide an overview of privacy risks associated with various sharing methods.
A summary of privacy attacks is presented in Table \ref{table:privacy_leakage}.

\begin{table*}[!t]
\centering
\footnotesize
\caption{Comparison of Privacy Attacks in FL}
\setlength\tabcolsep{2pt}
\begin{tabular}{c|cccccc|m{0.47\textwidth}}
\hline
\multirow{2}{*}{Ref} & \multicolumn{6}{c|}{Infer private information from}                                                                                                                                                         & \multicolumn{1}{c}{\multirow{2}{*}{Description}} \\ \cline{2-7}
                     & Gradients & \begin{tabular}[c]{@{}c@{}}Model\\ parameters\end{tabular} & \begin{tabular}[c]{@{}c@{}}Synthetic \\ samples\end{tabular} & \begin{tabular}[c]{@{}c@{}}Generative \\ models\end{tabular} & Logits & Features &                              \\ \hline
                     \cite{zhu2019deep}&  $\checkmark$         &                                                            &                                                           &                                                             &        &          &  Reconstruct the original data by optimizing a pair of dummy inputs and labels based on minimizing the distance between dummy and target gradients.
                         \\ \hline
                     \cite{zhao2020idlg}&    $\checkmark$       &                                                            &                                                           &                                                             &        &          &  Extract the ground-truth labels from the shared gradients and leverage labels to reconstruct the training data via optimizing a gradient matching objective.
                            \\ \hline
                     \cite{yin2021see}&    $\checkmark$       &                                                            &                                                           &                                                             &        &          &  Recover labels from final layer gradients and introduce a group consistency regularization term to reconstruct private images from a batch of data.
                            \\ \hline
                            
                     \cite{gupta2022recovering}&    $\checkmark$       &                                                            &                                                           &                                                             &        &          &  Recovering multiple sentences from gradients using beam search and token reordering policies.
                            \\ \hline
                     \cite{melis2019exploiting}&    $\checkmark$       &          $\checkmark$                                                   &                                                           &                                                             &        &          &  Infer the membership and properties of training data based on the gradients and the global model.
                            \\ \hline
                     \cite{fredrikson2014privacy}&          &   $\checkmark$                                                           &                                                           &                                                             &        &          &  Reconstruct the most likely examples corresponding to a specific target label by applying the maximum a posteriori principle to model the inversion process.
                            \\ \hline
                     \cite{zhang2020secret}&         &      $\checkmark$                                                        &                                                           &                                                              &        &          &  
                          Reconstruct private by using a public dataset to learn a distributional prior for guiding the inversion process. 
                          \\  \hline
                     \cite{zari2021efficient}&           &   $\checkmark$                                                          &                                                           &                                                             &          &          &  
                        Train an attack model based on the temporal evolution of the output score assigned to infer membership. \\ \hline
                     \cite{ganju2018property}&           &   $\checkmark$                                                          &                                                             &                                                             &         &          &  Train a meta-classifier based on shadow models for property inference attacks.
\\ \hline                               
                     \cite{hilprecht2019monte}&           &                                                            &   $\checkmark$                                                        &                                                             &        &          &             Infer membership by counting the number of neighboring samples of the target samples.                 \\ \hline
                     \cite{van2023membership}&           &                                                            &  $\checkmark$                                                         &                                                             &        &          &              Infer membership by incorporating density estimation to detect the overfitting of the generative model.         \\ \hline
                     \cite{hayes2019logan}&           &                                                            &    $\checkmark$                                                       &        $\checkmark$                                                     &        &          &   Estimate membership based on the output confidence of the discriminator.
                     \\ \hline
                     \cite{chen2020gan}&           &                                                            &     $\checkmark$                                                      &    $\checkmark$                                                         &        &          &   Measure the distance between the target sample and the nearest synthetic sample for membership inference                         
                     \\ \hline
                     \cite{zhang2023ideal} &           &                                                            &                                                           &                                                           &   $\checkmark$     &          &  Develop an IDEAL method to distill knowledge and generate data from logits.
                           \\ \hline
                     \cite{salem2019ml} &           &                                                            &                                                           &                                                             &   $\checkmark$     &          &   Infer the membership of samples based on the statistical measures of logits.
                           \\ \hline
                     \cite{shokri2017membership1}&           &                                                            &                                                           &                                                             &  $\checkmark$      &          &  Construct multiple shadow models to mimic the target model’s behavior for membership inference.
                            \\ \hline
                     \cite{zhang2021leakage}&           &                                                            &                                                           &                                                             &  $\checkmark$      &          &  Infer population-level properties in datasets via a few hundred inference queries.
                            \\ \hline
                    \cite{nasr2019comprehensive}&           &                                                            &                                                           &                                                             &   $\checkmark$     &  $\checkmark$        &  Train a classifier in a supervised manner to infer the membership by taking the logits or features as input.
                            \\ \hline
                     \cite{ding2020privacy}&           &                                                            &                                                           &                                                             &        &   $\checkmark$       &  Train a classifier by collecting pairs of features and privacy labels to predict privacy attributes from intermediate features.
                            \\ \hline
                     \cite{dosovitskiy2016inverting}&           &                                                            &                                                           &                                                             &        &   $\checkmark$       &  Invert visual features with an up-convolutional neural network for image reconstruction.
                            \\ \hline
                     \cite{he2019model} &           &                                                            &                                                           &                                                             &        &   $\checkmark$       &Find an inverse mapping from the target feature to input data such that the distance between the feature of this input and the target feature is minimized.
                            \\ \hline
\end{tabular}
\label{table:privacy_leakage}
\end{table*}

\subsection{Privacy Leakage in Model Sharing}

Sharing gradients or model parameters in FL makes the system vulnerable to white-box attacks, which enables adversaries to reconstruct the private training data and uncover the properties of the dataset.

\subsubsection{Inferring Private Information from Gradients}
Gradients can leak sensitive information about local data, such as original data and private properties \cite{zhu2019deep, zhao2020idlg, geiping2020inverting, yin2021see}.
For instance, deep leakage from gradients (DLG) \cite{zhu2019deep} exploits the private information contained in the gradients, which optimizes a pair of dummy inputs and labels to minimize the distance between dummy and target gradients, thus recovering the original data.  
Later, Zhao et al. \cite{zhao2020idlg} improved DLG by incorporating the signs of gradients to reveal the ground-truth labels.
Yin et al. \cite{yin2021see} proposed GradInversion to reconstruct more realistic images by introducing batch label restoration.
For textual data, Balunovic et al. \cite{balunovic2022lamp}  introduced LAMP to leak the training data by utilizing cosine similarity and embedding regularization terms in conjunction with the gradient matching loss.
A recent study by Gupta et al. \cite{gupta2022recovering} demonstrated the possibility of recovering multiple sentences from gradients using beam search and token reordering policies.
Besides, authors of \cite{melis2019exploiting} utilized the non-zero gradients of the embedding layer to reveal the words in the training batch.
They also trained a batch property classifier to infer whether the data used to calculate the gradient have a specific property or not.

\subsubsection{Inferring Private Information from Model Parameters}

Early studies \cite{fredrikson2014privacy,fredrikson2015model} pointed out that linear models are vulnerable to data reconstruction attacks.
Subsequently, Fredrikson et al. \cite{fredrikson2015model} applied the maximum a posteriori principle to reconstruct the most likely examples corresponding to a specific target label.
In addition, authors of \cite{zhang2020secret} proposed a novel GAN-based technique called generative model-inversion (GMI) that utilizes a public dataset to learn a distributional prior for guiding the inversion process, instead of reconstructing private data from scratch.
Moreover, model parameters can be exploited in a white-box setting to infer both membership and private properties of datasets.
For membership inference,  Leino et al. \cite{leino2020stolen}  analyzed the model's idiosyncratic use of features for white-box membership inference.
Similarly, authors of \cite{zari2021efficient} proposed to passively infer membership based on the predictions of local models across different epochs.
To determine whether a model possesses a specific property, a meta-classifier was trained in \cite{ateniese2015hacking}. 
Specifically, they collected several training sets with or without a specific property and used them to train different models.
The parameters of these models are subsequently employed to train a meta-classifier for property inference.
Ganju et al. \cite{ganju2018property} later extended the approach to fully connected neural networks.
They arranged the neural networks into a canonical form and represented each neural network layer as a set, thus reducing the complexity of the meta-classifier.

\subsection{Privacy Leakage in Synthetic Data Sharing}

When sharing synthetic data or generative models among clients in FL, attackers can infer private information from synthetic samples in a black-box manner or from generative models in a white-box manner.

\subsubsection{Inferring Private Information from Synthetic Data}
Authors of \cite{hayes2019logan} trained a GAN model to approximate the distribution of synthetic samples.
The discriminator of this GAN can be utilized to perform the membership inference attacks.
Specifically, the data samples that have high output probabilities are classified as private training samples.
Besides, Chen et al. \cite{chen2020gan} utilized the distance between the target sample and the nearest synthetic sample as the metric for membership inference.
A smaller distance implies a higher probability that the target sample is from the private training set.
Similarly, Hilprecht et al. \cite{hilprecht2019monte} used Monte Carlo integration \cite{shapiro2003monte} to calculate the number of close neighborhood samples as the metric to infer the membership.
Moreover, DOMIAS \cite{van2023membership} releases the black-box setting by including a reference dataset that has the same distribution but different identities as the training dataset.
A density-based attack method is proposed to infer membership by targeting local overfitting of the generative model. 
Specifically, a target sample is considered part of the training dataset if it has a higher probability of matching generated samples than those in the reference dataset.

\subsubsection{Inferring Privacy from Generative Models}

In FL studies that involve collaborative training of generative models among clients, the sharing of generative models becomes necessary.
Compared with synthetic data sharing, the generator and the discriminator in generative models are vulnerable to inference attacks due to their potential to memorize additional private information.
If the discriminators are shared among clients, they can be used for membership inference, as each discriminator is optimized for discriminating between real and synthetic samples \cite{hayes2019logan}.
The trained discriminator is more likely to classify samples from the training dataset as real.
With access to the generative models, Chen et al. \cite{chen2020gan} proposed to optimize the input latent features of the generator to obtain the nearest synthetic sample to the target sample.
A precise measure of the distance between the target sample and the synthetic sample enables an accurate estimation of the probability for membership inference.

\subsection{Privacy Leakage in Knowledge Sharing}

While sharing knowledge in FL largely reduces the amount of information exposed to other parties, the logits or intermediate features are still vulnerable to privacy attacks.

\subsubsection{Inferring Private Information from Logits}

Previous studies \cite{li2022swing,zhang2023ideal} have empirically demonstrated that sharing logits can inadvertently lead to the disclosure of sensitive information.  
In \cite{zhang2023ideal}, the authors proposed the IDEAL method, which query-efficiently reconstructs the private training samples from logits.
Besides, deep learning models are susceptible to memorizing and overfitting the training dataset, leading to different behaviors on the training data (members) and test data (non-members).
Such vulnerability enables the attacker to infer the membership status of a data record from logits.
Specifically, the studies \cite{salem2019ml, yeom2018privacy} determined membership status based on various metrics, such as accuracy, prediction loss, confidence scores, and output entropy. 
Moreover, the method proposed in \cite{nasr2019comprehensive} involves training a binary classifier in a supervised manner, which distinguishes between member and non-member samples based on their logits.
However, in most cases, an attacker may not have access to the private member data. 
The authors of \cite{shokri2017membership1} developed an unsupervised membership inference attack, which creates multiple shadow models to mimic the behavior of the target model. 
The attacker gains access to the training and test datasets of these shadow models, enabling them to create a new dataset containing features and ground truth of membership.
In addition, the authors of \cite{zhang2021leakage} studied the property inference attacks when only black-box access to the model is available to the attacker.
They found that making only a few hundred queries to the model is sufficient to infer population-level properties within datasets.

\subsubsection{Inferring Private Information from Intermediate Features}

The authors of \cite{ding2020privacy} leveraged the intermediate features to predict privacy properties that are unrelated to the target task.
Once there is a dataset containing auxiliary samples with privacy labels, the privacy attacker can establish a new neural network to learn the mapping from the feature vector to the privacy properties.
Similarly, Nasr et al. \cite{nasr2019comprehensive} utilized the activations of intermediate layers to infer the membership of data points.
Dosovitskiy et al.\cite{dosovitskiy2016inverting} inverted image feature vectors with up-convolutional networks.
This neural network is trained to predict the \emph{expected pre-image} according to a given feature vector.
Such an expected pre-image represents the weighted average of all possible images that produce the given feature vector. 
Besides, the authors of \cite{tuor2018understanding} recovered the raw data by using the output of the intermediate layers. 
The main idea is to find an image such that the error between the feature vector of this image and the target feature is the smallest.
Additionally, Inverse-Network  \cite{he2019model} identifies the mapping from the intermediate features to the input by querying the black-box model and then approximates the inverse function to reconstruct the private samples.

\subsection{Summary of Privacy Leakage}

This section surveys various privacy attacks in FL, which can be broadly classified into three types: data reconstruction attacks, membership inference attacks, and property inference attacks.
Data reconstruction attacks aim to extract the raw private data from the shared information. Membership inference attacks, on the other hand, attempt to infer whether a specific data sample is used to train the local model. Similarly, property inference attacks tend to infer specific properties of private datasets.
Overall, all sharing methods are susceptible to these privacy attacks. 
Sharing model updates is vulnerable to white-box attacks, where adversaries can reconstruct private data through gradient inversion or model inversion. 
Without exposing the model parameters, sharing synthetic data or knowledge is immune to white-box attacks.
However, these sharing methods still face potential privacy risks, where the private information encoded within synthetic samples or logits can be exploited for membership inference and property inference.
To counteract these privacy attacks, it is crucial to develop robust defense mechanisms.

\section{Defense against Privacy Leakage}

\label{sec:defense}

\begin{figure}[t]
\centering
\includegraphics[width=0.85\linewidth]{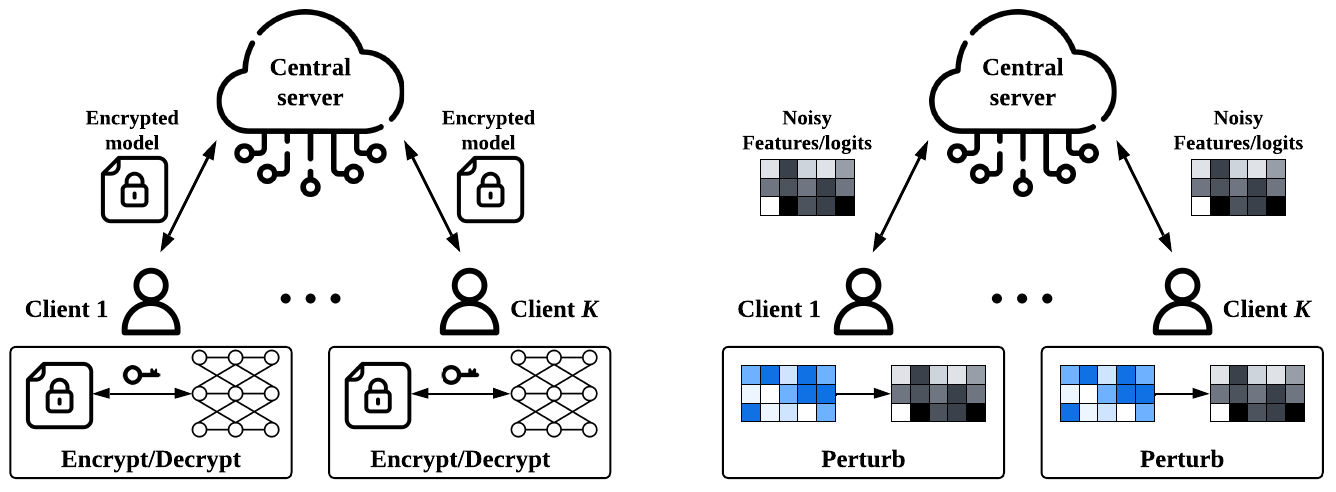}
\caption{Defense methods against privacy attacks in FL: (Left) cryptography-based technique for model sharing and (Right) perturbation method for knowledge sharing.}
\label{Fig:two_types_of_defense_methods}
\end{figure}

To enhance privacy guarantees in FL, two main strategies are cryptography-based techniques and perturbation methods, as depicted in Fig. \ref{Fig:two_types_of_defense_methods}.
Cryptographic primitives enable the computation over encrypted data, and perturbation methods add noise during the federated training.
In this section, we delve into various defense strategies designed for different sharing methods.
A summary of these studies is available in Table \ref{table:defense_methods}.

\subsection{Defense Methods for Model Sharing}

\subsubsection{Secure Model Aggregation}

Secure model aggregation is a class of cryptographic protocols that can privately aggregate models from multiple clients without revealing any individual models.
The first secure aggregation method was proposed in \cite{bonawitz2017practical}.
The key idea is to adopt pairwise masks to conceal models.
These masks have an additive structure that can be canceled after aggregation.
Later, Bonawitz et al. \cite{bonawitz2019federated} leveraged an aggressive quantization approach to develop a communication-efficient secure aggregation scheme. 
Another line of cryptography-based techniques for secure aggregation is homomorphic encryption (HE), which allows performing arithmetic computations directly on encrypted data without decryption.
Jin et al. \cite{jin2023fedml} presented a practical federated learning system with efficient HE-based secure model aggregation, namely FedML-HE, which selectively encrypts sensitive parameters.
Recently, the authors of \cite{pedrouzo2023practical}  introduced a lightweight communication-efficient multi-key approach for secure model aggregation, which combines the secret-key RLWE-based HE \cite{lyubashevsky2010ideal} and additive secret sharing.

\begin{table*}[!t]
\centering
\footnotesize
\caption{Comparison of Defense Strategies}
\setlength\tabcolsep{2pt}
\begin{tabular}{c|cccccc|m{0.47\textwidth}}
\hline
\multirow{2}{*}{Ref} & \multicolumn{6}{c|}{Provide privacy protection on}                                                                                                                                                         & \multicolumn{1}{c}{\multirow{2}{*}{Description}} \\ \cline{2-7}
                     & Gradients & \begin{tabular}[c]{@{}c@{}}Model\\ parameters\end{tabular} & \begin{tabular}[c]{@{}c@{}}Synthetic \\ samples\end{tabular} & \begin{tabular}[c]{@{}c@{}}Generative \\ models\end{tabular} & Logits & Features &                              \\ \hline
                     \cite{yadav2020differential}  &    $\checkmark$       &                                                          &                                                           &                                                             &        &          &     Use Gaussian and Laplace mechanisms to provide a DP guarantee to the gradients.                       \\ \hline
                     \cite{wei2021gradient}  &    $\checkmark$       &                                                           &                                                           &                                                             &        &          &   Present a gradient leakage resilient approach based on a dynamic decay noise-injection policy.             \\ \hline
                     \cite{zhu2019deep}   &    $\checkmark$       &                                                           &                                                           &                                                             &        &          &     Prune gradients with small magnitude to zero to reduce the information leakage.                        \\ \hline
                     \cite{fan2020rethinking}   &    $\checkmark$       &                                                           &                                                           &                                                             &        &          &     Develop a secret polarization network to perturb the gradients adaptively.                        \\ \hline
                     \cite{bonawitz2017practical}&         $\checkmark$   &  $\checkmark$                                                          &                                                           &                                                             &        &          &   Use secure multiparty computation protocols to protect gradients and model parameters.     \\ \hline
                     \cite{bonawitz2019federated}&   $\checkmark$        &    $\checkmark$                                                        &                                                           &                                                             &        &          &   Leverage the aggressive quantization approach to build a communication-efficient secure aggregation system.                           \\ \hline
                     \cite{jin2023fedml} &  $\checkmark$       &       $\checkmark$                                                     &                                                           &                                                             &        &          &     Build a practical FL system with efficient HE-based secure model aggregation.                        \\ \hline
                     \cite{pedrouzo2023practical}  &   $\checkmark$       &       $\checkmark$                                                     &                                                           &                                                             &        &          &    Propose a communication-efficient multi-key approach for model aggregation by combining RLWE-based HE and additive secret sharing.                       \\ \hline
                     
                     \cite{han2019logistic}&           &                                                            &  $\checkmark$                                                         &                                                             &        &          & Implement HE on machine learning problems by approximating objective functions with polynomials.                            \\ \hline
                     \cite{dres-fl}&           &                                                            &               $\checkmark$                                            &                                                             &        &          &    Adopt Lagrange coding to secretly share  samples.                        \\ \hline
                     \cite{sun2022stochastic2}&           &                                                            &          $\checkmark$                                                 &                                                             &        &          &            Adopt coded computing and add Gaussian noise to protect the privacy of shared data.                 \\ \hline
                     \cite{lee2021digestive}&           &                                                            &                                       $\checkmark$                    &                                                             &        &          & Generate distorted data using a digestive neural network for privacy protection.                       \\ \hline
                     \cite{xie2018differentially}&           &                                                            &                                                           &                          $\checkmark$                                    &       &          &  Propose a DPGAN model to achieve DP in GANs by adding carefully designed noise to gradients.
                         \\ \hline
                     \cite{jordon2018pate}&           &                                                            &                                                           &                       $\checkmark$                                      &        &          &     Enable DP by leveraging the PATE framework to inject noise into predictions.                        \\ \hline
                     \cite{augenstein2019generative}&           &                                                            &                                                           &              $\checkmark$                                               &        &          &                   Achieve user-level DP by clipping per-user updates.  \\ \hline
                     \cite{hamm2016learning}&           &                                                            &                                                           &                                                             & $\checkmark$       &          & Minimize the regularized weighted empirical risk with soft labels and perturb the model outputs.
                             \\ \hline
                     \cite{papernotsemi} &           &                                                            &                                                           &                                                             &   $\checkmark$     &          &   Adopt noisy voting to aggregate predictions from local models, where vote counts are perturbed by Laplacian noise.
                           \\ \hline
                     \cite{lyu2020towards} &           &                                                            &                                                           &                                                             & $\checkmark$       &          &  Adopt the improved Binomial Mechanism and Discrete Gaussian Mechanism to achieve distributed DP, and utilize HE for secure aggregation.
                            \\ \hline
                     \cite{li2022swing} &           &                                                            &                                                           &                                                             &  $\checkmark$      &          &  Protect the private information of local models by dynamically and adaptively adjusting the temperature coefficient according to the degree of private information contained in the data.
                            \\ \hline
                     \cite{qi2023differentially} &           &                                                            &                                                           &                                                             & $\checkmark$       &         &    Actively use small, publicly-selected datasets to transfer knowledge and perturb the predictions via a randomized response mechanism.
                          \\ \hline
                     \cite{gong2022preserving}&           &                                                            &                                                           &                                                             & $\checkmark$       &          &   Quantize and add noise on logits of models for aggregation and distillation.
                           \\ \hline
                     \cite{sun2020federated}&           &                                                            &                                                           &                                                             & $\checkmark$       &          & Provide a noise-free DP guarantee in federated distillation via random data sampling.
                             \\ \hline
                     \cite{mai2023split}&           &                                                            &                                                           &                                                             &        &  $\checkmark$        & Introduce Split-N-Denoise that perturbs the features by the addition of Laplacian noise.
                             \\ \hline
                     \cite{xue2021dp}&           &                                                            &                                                           &                                                             &        &  $\checkmark$        &  Achieve DP-Image by adding noise to the image feature vector.
                             \\ \hline
\end{tabular}
\label{table:defense_methods}
\end{table*}

\subsubsection{Model Perturbation}
While secure model aggregation can protect local models against attacks, it is still possible for attackers to reveal private information from the aggregated model.
To address this issue, model perturbation methods \cite{abadi2016deep,zheng2021federated,bu2020deep,fan2020rethinking,sun2021soteria,zhang2021matrix} were proposed to provide stronger privacy guarantees. 
Among these methods, differential privacy (DP) \cite{dwork2006differential} enables an untrusted central collector to perform privacy-preserving data analytics.
The differentially private SGD algorithm proposed in \cite{abadi2016deep} is one of the main client-side defense methods, using a privacy accountant to constrain total privacy loss. 
Subsequently, several studies \cite{yadav2020differential, hao2019towards,wei2021gradient} have integrated DP with FL to protect local models against privacy attacks.
For example, Yadav et al. \cite{yadav2020differential} used Gaussian and Laplace mechanisms to secure updated gradients in each communication round. 
Wei et al. \cite{wei2021gradient} proposed a dynamic decay noise-injection policy to improve the gradient-leakage resilience.
In addition to the DP mechanism, Fan et al. \cite{fan2020rethinking} developed a secret polarization network to adaptively add noise to gradients, and Zhu et al. \cite{zhu2019deep} adopted gradient pruning to defend against gradient inversion attacks.

\subsection{Defense Methods for Synthetic Data Sharing}

\subsubsection{Synthetic Data Encryption}

Encryption is a crucial method for safeguarding sensitive information and ensuring its confidentiality.
For instance, HE techniques enable computations to be performed on encrypted data without the need for decryption \cite{paillier1999public,brakerski2012fully}. 
This feature is particularly useful for encrypting synthetic data in FL. To enhance the computational efficiency of HE, Graepel et al. \cite{graepel2013ml} proposed to approximate objective functions using polynomials for binary classification problems.
Expanding on this idea, the studies in \cite{kim2018logistic,han2019logistic} propose approximate HE schemes for conducting logistic regression on homomorphically encrypted data. 
Besides, the authors of \cite{hesamifard2017cryptodl,gilad2016cryptonets} focused on training neural networks using encrypted data, while the authors of \cite{tastan2023capride,afonin2021towards} performed distillation based on encrypted samples. 
In addition to HE protocols, secret data sharing techniques can also protect the privacy of individual clients while enabling data sharing and aggregation.
The studies in \cite{dres-fl, so2021turbo, so2020scalable} employ Lagrange coding to securely share data in FL, 
allowing the computation of model updates on the encoded data.

\subsubsection{Privacy-Preserving Data Generation}
To defend against attacks targeting synthetic data, one simple solution is to directly perturb the generated data.
For instance, coded federated learning \cite{sun2022stochastic,sun2022stochastic2,anand2021differentially} perturbs the random linear combination of the private data by adding Gaussian noise.
Besides, the digestive neural network \cite{lee2021digestive} modifies data to digested data batches, which maintains the classification performance and alleviates privacy leakage.
To generate differentially private synthetic samples, DPGAN \cite{xie2018differentially} incorporates differentially private stochastic gradient descent \cite{song2013stochastic} into the GAN training process, and PATE-GAN \cite{jordon2018pate} applies the Private Aggregation of Teacher Ensembles (PATE) framework \cite{pate} to the GAN model.
Recently, DP-FedAvg-GAN \cite{augenstein2019generative} develops DP in FL, where users clip the local updates to maintain user-level DP guarantees.

\subsection{Defense Methods for Knowledge Sharing}

\subsubsection{Private Knowledge Aggregation}

The simplest approach for privately aggregating logits from auxiliary samples is majority voting.
However, Hamm et al. \cite{hamm2016learning} showed that such a mechanism is sensitive to individual votes, rendering the aggregated knowledge vulnerable to privacy attacks. 
To address this issue, they proposed a new empirical risk, where each sample is weighted by the confidence of the ensemble.
Later, Papernot et al. \cite{papernotsemi} proposed the PATE approach, where the knowledge for distillation is aggregated through noisy voting among all local models.
Recently, the authors of \cite{zhao2022privacy} improved the PATE framework by combining secret sharing with Intel Software Guard Extensions (SGX) \cite{mckeen2016intel}.
Additionally, Lyu et al. \cite{lyu2020towards} combined distributed differential privacy (DDP) and HE to ensure that the aggregator learns nothing but the noisy aggregated prediction.

\subsubsection{Knowledge Perturbation}

Another research avenue providing privacy protection for knowledge dissemination involves developing perturbation techniques.
In federated knowledge distillation, Li et al. \cite{li2022swing} developed a swing distillation technique to adaptively assign different temperatures to tokens based on the likelihood that a token in a position contains private information.
Besides, the authors of \cite{qi2023differentially} proposed a differentially private knowledge transfer framework, where the clients perturb local predictions via a randomized response mechanism.
Similarly, the studies \cite{mai2023split,xue2021dp} add noise to the intermediate features to provide DP guarantee, and Gong et al. \cite{gong2022preserving} employed quantization on the locally-computed logits to enhance privacy guarantees. 
However, directly adding noise or quantizing local predictions may introduce a substantial tradeoff between privacy budget and model performance.
The FedMD-NFDP method proposed in \cite{sun2020federated} applies a Noise-Free Differential Privacy (NFDP) mechanism with random sampling to guarantee the privacy of each model.

\subsection{Summary of Defense Methods}

This section provides an overview of defense methods against privacy attacks.
The existing studies can be broadly classified into two major types: cryptography-based techniques and perturbation methods. 
Cryptographic primitives enable the computation of models, synthetic data, or knowledge in an encrypted form without privacy leakage.
Perturbation methods focus on adding noise to privacy-sensitive information.
DP mechanisms are the most common techniques within this category, which can add randomness to different types of shared information and mathematically quantify the privacy guarantees.
However, both cryptography-based and perturbation methods have their weaknesses. 
Cryptography-based techniques offer robust security, but they generally introduce significant computational and communication overhead \cite{bonawitz2017practical, jahani2022swiftagg}.
Perturbation methods, while generally more efficient, require careful tuning of noise levels to balance privacy and utility.
This is because excessive noise obscures crucial information and thus compromises training performance \cite{kim2021federated,yadav2020differential}.

\section{Experiments}

\label{sec:exp}

The objective of the experimental study is to analyze the performance of different sharing methods in FL in terms of accuracy and communication overhead.
In addition, we analyze privacy leakage under various attacks and evaluate the effectiveness of the defense methods.

\subsection{Setup}

\subsubsection{Datasets} We select two benchmark datasets for the experiments: SVHN \cite{svhn} and CIFAR-10 \cite{krizhevsky2009learning}.
To simulate the non-IID distribution, we set the number of participating clients to 10 and adopt the Dirichlet distribution to shuffle the training data.
Specifically, the parameter $\alpha$ of the Dirichlet distribution controls the data heterogeneity, which is selected from $\{0.01, 0.1\}$ on SVHN and $\{0.05, 0.1\}$ on CIFAR-10.

\subsubsection{FL Methods}

Based on the taxonomy of various sharing schemes, we choose the following representative methods from each category for our experimental analysis.

\begin{itemize}
\item \textbf{Model sharing}: 
We evaluate the performance of three model sharing methods, namely FedAvg~\cite{fedavg}, FedProx~\cite{fedprox}, and FedPAQ~\cite{reisizadeh2020fedpaq}.
FedProx is a variant of FedAvg that introduces a proximal term to tackle heterogeneity.
In FedPQA, the clients quantize every floating-point value in the model updates to a 4-bit representation to reduce the communication overhead.
\item \textbf{Synthetic data sharing}:
We evaluate two synthetic data sharing methods: dataset condensation (DC) \cite{dataset_condensation} and Mixup \cite{zhangmixup}.
The DC method condenses a large dataset into a small set of informative synthetic samples, while Mixup performs a linear interpolation between local samples to generate augmented samples.
    
\item \textbf{Knowledge sharing}: We evaluate FedMD \cite{li2019fedmd}, FedED \cite{sui2020feded}, and FKD \cite{jeong2018communication_FKD1,seo202216_FKD2} in the experiments. 
FKD is a data-free knowledge distillation method that introduces a distillation regularizer to penalize the gap between the clients' class-wise predictions and the ensemble predictions.
In contrast, FedMD and FedED rely on a proxy dataset to 
transfer knowledge. 
Clients in FedMD use the aggregated logits to fine-tune the local models, while FedED trains an extra model at the server based on the local predictions.
The proxy data used to transfer knowledge in the SVHN and CIFAR-10 classification tasks are from MNIST \cite{baldominos2019survey} and Tiny ImageNet \cite{le2015tiny_imagenet} datasets, respectively.

\end{itemize}

In addition, we also evaluate the performance of FedMix~\cite{fedmix}, which shares both model parameters and synthetic samples generated by the mixup technique.
Moreover, we present two accuracy bounds for comparisons.
Firstly, we report the average accuracy of local models by independently training on their respective local datasets (referred to as IndepTrain), which serves as a lower bound in the comparison.
Secondly, we report the accuracy achieved by centralized training, which represents the performance upper bound.
Table \ref{table:methods_attacks} summarizes the methods evaluated in experiments.

\begin{table*}[t]
\centering
\footnotesize
\caption{Representative Methods Evaluated in Experiments}
\begin{tabular}{c|ccc}
\hline
\multirow{2}{*}{Method} & \multirow{2}{*}{Type}           & \multicolumn{2}{c}{Privacy attacks}                   \\ \cline{3-4} 
                        &                                 & Model inversion attack & Membership inference attack \\ \hline
FedAvg \cite{fedavg}                  & Model sharing                   & White-box              & White-box                   \\
FedProx \cite{fedprox}                 & Model sharing                   & White-box              & White-box                   \\
FedPAQ \cite{reisizadeh2020fedpaq}                  & Model sharing                   & White-box              & White-box                   \\
FedMix \cite{fedmix}                  & Model \& Synthetic data sharing & White-box              & White-box                   \\
DC \cite{dataset_condensation}                     & Synthetic data sharing          & --                     & --                          \\
Mixup \cite{zhangmixup}                   & 
Synthetic data sharing          & --                     & --                          \\
FedMD \cite{li2019fedmd}                   & Knowledge sharing               & Black-box              & Black-box                   \\
FedED \cite{sui2020feded}                   & Knowledge sharing               & Black-box              & Black-box                   \\
FKD \cite{jeong2018communication_FKD1,seo202216_FKD2}                    & Knowledge sharing               & --                     & --                          \\\hline
IndepTrain                     & --          & --                     & --                          \\
Centralized                     & Raw data sharing          & --                     & --                          \\ \hline
\end{tabular}
\label{table:methods_attacks}
\end{table*}

\subsubsection{Privacy Attacks}

Our analysis primarily focuses on the most common privacy attacks in the literature, including model inversion attacks and membership inference attacks.
In particular, we pay close attention to server-side attacks, as the server can acquire more information than individual clients, thereby posing the most significant threat to privacy.

\begin{itemize}
\item \textbf{Model inversion attack}: 
Following \cite{zhang2020secret, zhang2023ideal}, we employ a GAN \cite{creswell2018generative} to extract knowledge from the shared information.
For the model sharing methods, the attacker can obtain each local model and directly use it as the discriminator to train the generator in GAN.
For knowledge sharing methods where the attacker can only access the models' output, we implement the IDEAL approach \cite{zhang2023ideal} to reconstruct private data in a black-box setting.
In the experiments, we evaluate model inversion attacks using the Frechet inception distance (FID) score \cite{heusel2017gans_FID} as a metric, which compares the distribution of the reconstructed images to the distribution of real images.
A lower FID value indicates better image quality and diversity.

\item \textbf{Membership inference attack}: 
The attacker employs a neural-based binary classifier to determine the membership of samples.
Following \cite{salem2019ml,rezaei2021difficulty}, we provide the attacker with more advantages than previous works \cite{shokri2017membership1,salem2019ml}.
Specifically, for the knowledge sharing methods, we assume the attacker can query the model, own a training dataset with the membership status, and observe the shared information during training.
For the model sharing methods, the attacker can further utilize the intermediate features and gradients for membership inference.
To evaluate the performance of membership inference, we create a membership dataset that comprises 5,000 member samples and 5,000 non-member samples randomly selected from the local datasets and the test dataset, respectively, and we use binary classification accuracy as the performance metric. 

\end{itemize}

To the best of our knowledge, no prior research has been conducted on performing model inversion attacks solely using shared data or class-wise predictions.
Thus, we assume that the attackers randomly guess the results when performing membership inference attacks. 
In addition, when conducting model inversion attacks on DC and Mixup, and on FKD and IndepTrain, we regard the synthetic data as the reconstructed samples for the former, and random noise as the reconstructed samples for the latter.

\subsubsection{Neural Network Architecture}

For the image classification task on SVHN and CIFAR-10 datasets, we employ a convolutional neural network (CNN) that comprises two convolutional layers with a kernel size of $5 \times 5$, two max-pooling layers with a kernel size of $2 \times 2$, and two fully-connected layers. 
The output channels in the convolutional layers are 64 and 128, respectively.
The input neurons of the fully-connected layers are 1600 and 512, respectively.
The generator in GAN for private data reconstruction has two fully-connected layers and two transposed convolution layers.
The binary classifier for membership inference comprises two fully-connected layers, where the number of hidden neurons is 128.

\begin{figure*}[t]
\centering
\subfigure[$\alpha = 0.01$]{
\centering
\includegraphics[width=0.98\linewidth]{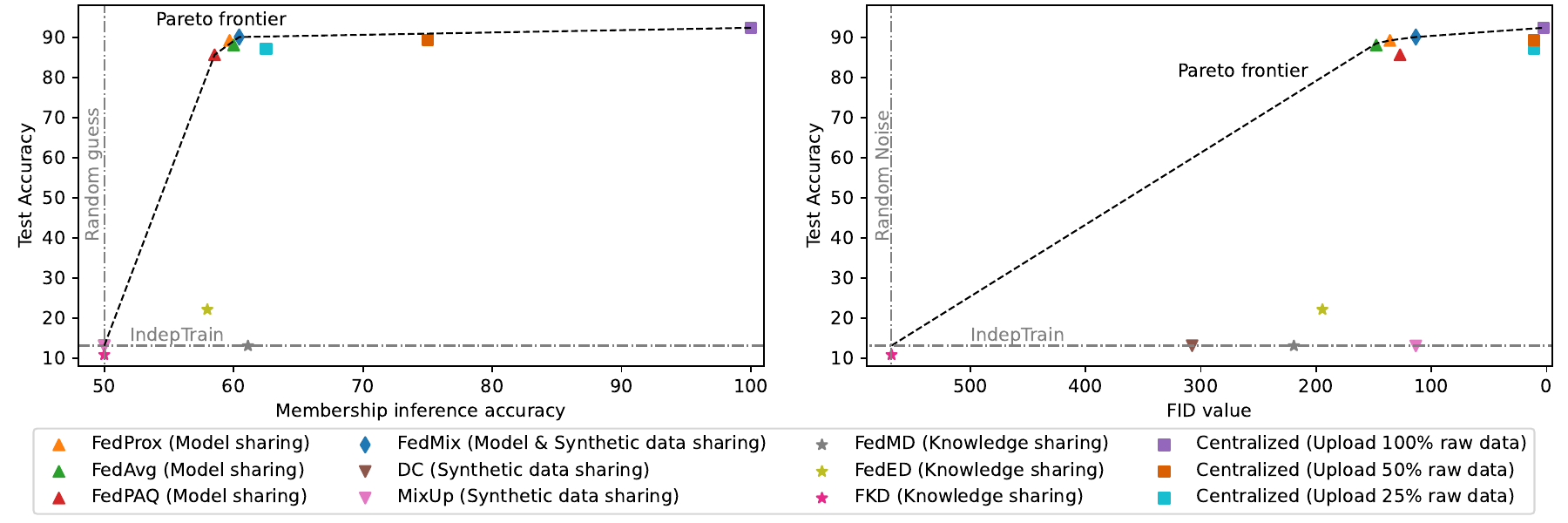}
}
\vfill
\subfigure[$\alpha = 0.1$]{
\centering
\includegraphics[width=0.98\linewidth]{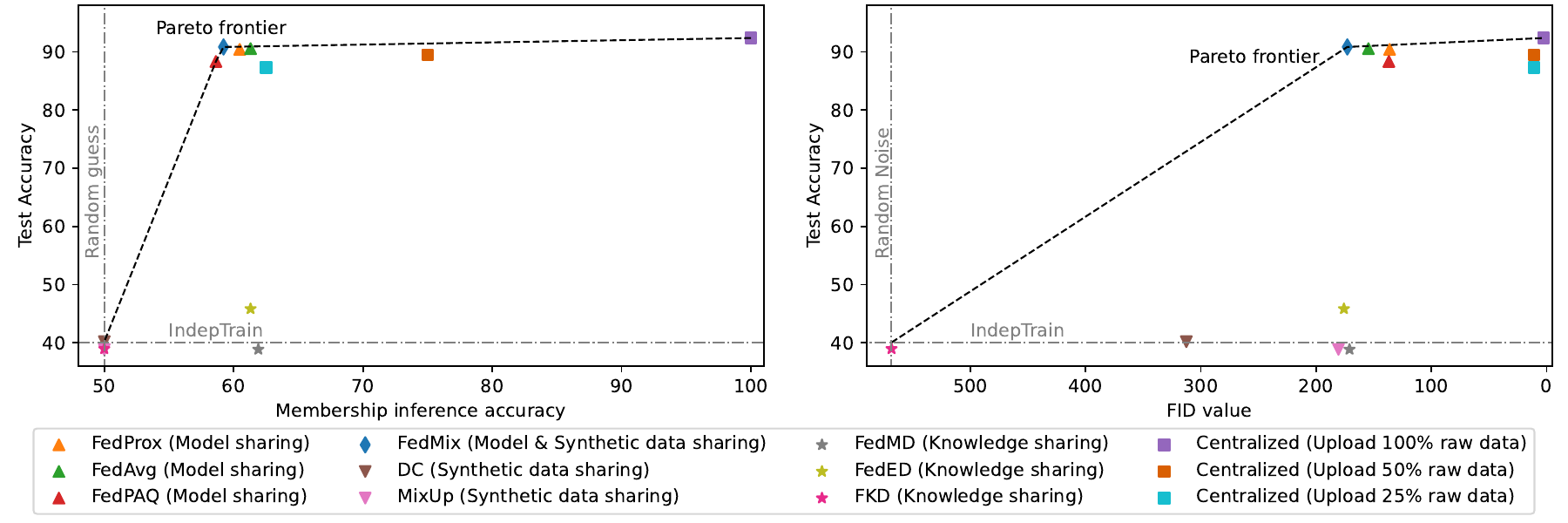}
}

\caption{Accuracy and privacy leakage in SVHN image classification task. Local datasets follow Dirichlet distribution with (a) $\alpha = 0.01$ and (b) $\alpha = 0.1$.}
\label{Fig:SVHN_results_tradeoff}
\end{figure*}

\begin{figure*}[t]
\centering
\subfigure[$\alpha = 0.05$]{
\centering
\includegraphics[width=0.98\linewidth]{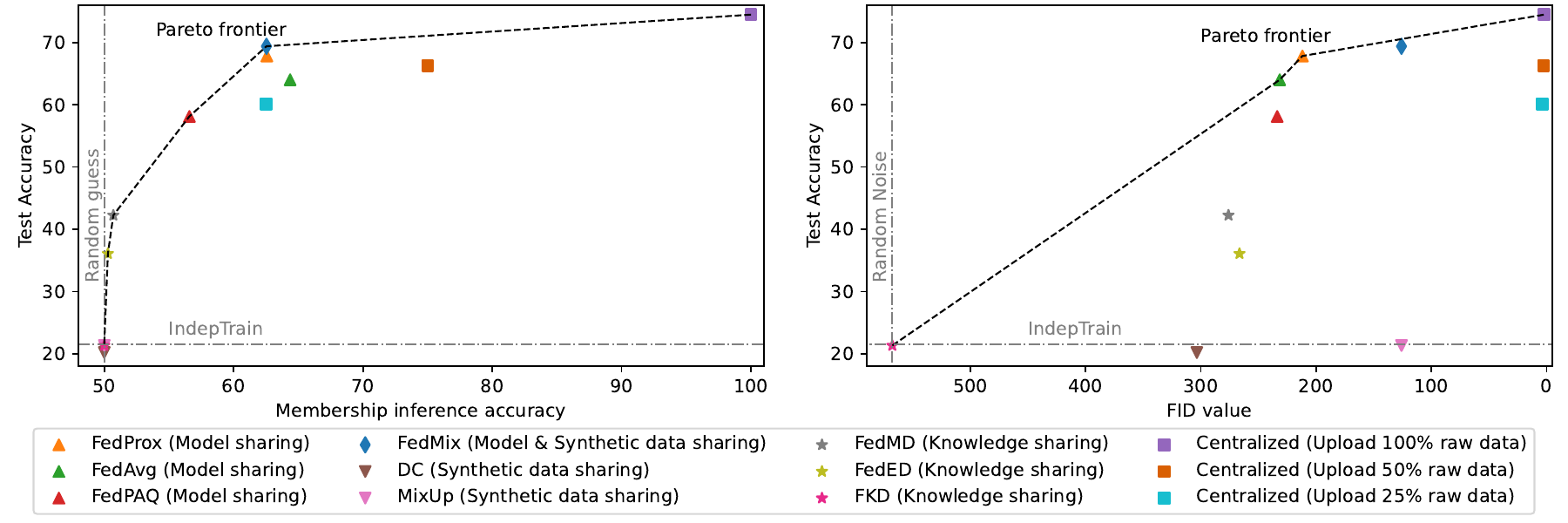}
}
\vfill
\subfigure[$\alpha = 0.1$]{
\centering
\includegraphics[width=0.98\linewidth]{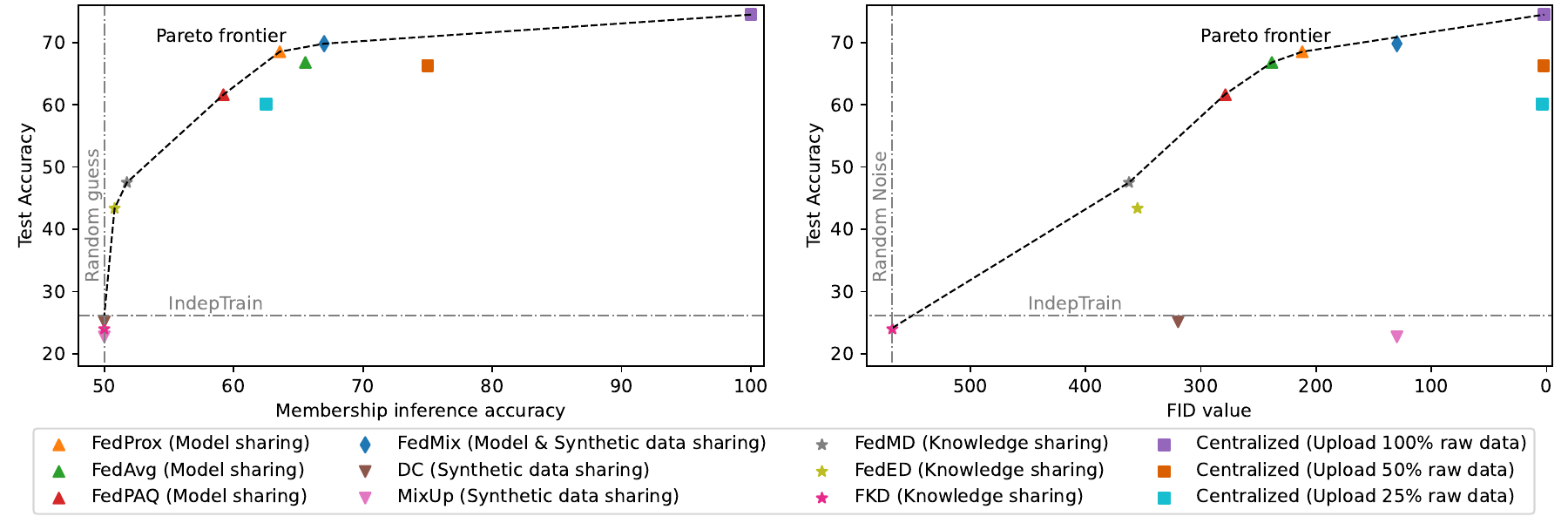}
}
\caption{Accuracy and privacy leakage in CIFAR-10 image classification task. Local datasets follow Dirichlet distribution with (a) $\alpha = 0.05$ and (b) $\alpha = 0.1$.}
\label{Fig:cifar10_results_tradeoff}
\end{figure*}

\subsection{Privacy-Utility Tradeoff}

\label{sec:subsec:exp_privacy_utility}

In this subsection, we evaluate the accuracy of various methods and investigate their privacy leakage.
We assume that the semi-honest server conducts privacy attacks by utilizing the shared information from clients.
We set the number of communication rounds to 1,000, with each round consisting of 10 local training steps.
For knowledge sharing methods, we set the distillation steps per communication round to 50.
The empirical results are presented in Fig. \ref{Fig:SVHN_results_tradeoff} and Fig. \ref{Fig:cifar10_results_tradeoff}, which report the classification accuracy, membership inference accuracy, and FID value of each method on the privacy-utility planes. Moreover, we plot the Pareto optimal curve to depict the tradeoff between performance and privacy leakage.
It is observed that with the increase of the classification accuracy, the membership inference accuracy approaches 100\%, and the FID value approaches zero.
Pursuing superior model performance can inadvertently lead to substantial compromises in data privacy.

Although centralized training achieves the highest test accuracy, it breaches the privacy requirement since local datasets of clients need to be collected.
Among various FL baselines, model sharing methods achieve comparable accuracy to the centralized training scheme.
In particular, FedPAQ achieves lower accuracy and privacy leakage compared with other model sharing methods, which is attributed to the reduced amount of shared information after quantization.
Furthermore, FedMix achieves additional performance gains by sharing synthetic data. 
However, this method also leads to relatively high privacy leakage due to increased exposure of clients' information.

The synthetic data sharing methods, Mixup and DC, demonstrate suboptimal results with respect to both utility and privacy guarantees.
This may be due to the synthetic data that cannot capture the underlying distribution of the real dataset.
The Mixup method generates new training examples by linearly interpolating input samples and their corresponding labels. 
However, this technique falls short when the mapping from input samples to their labels is non-linear.
For the DC method, as the condensed dataset may contain numerous task-irrelevant features \cite{lee2022dataset}, this issue potentially leads to a degradation in model performance.

The knowledge sharing methods offer much stronger privacy guarantees compared with the model sharing methods.
By keeping the model parameters locally, the local model is protected from white-box privacy attacks, resulting in lower membership inference accuracy and larger FID value.
Moreover, the data-based knowledge sharing methods, such as FedMD and FedED, outperform the data-free knowledge sharing method, FKD, in terms of test accuracy.
This is because the knowledge shared in a data-free manner could be misleading and ambiguous \cite{wang2022knowledge,shao2023selective}.
On the other hand, FKD demonstrates greater robustness against privacy attacks than both FedMD and FedED due to the reduced exposure of information during the training process.

\begin{figure*}[t]
    \centering
    \includegraphics[width=\textwidth]{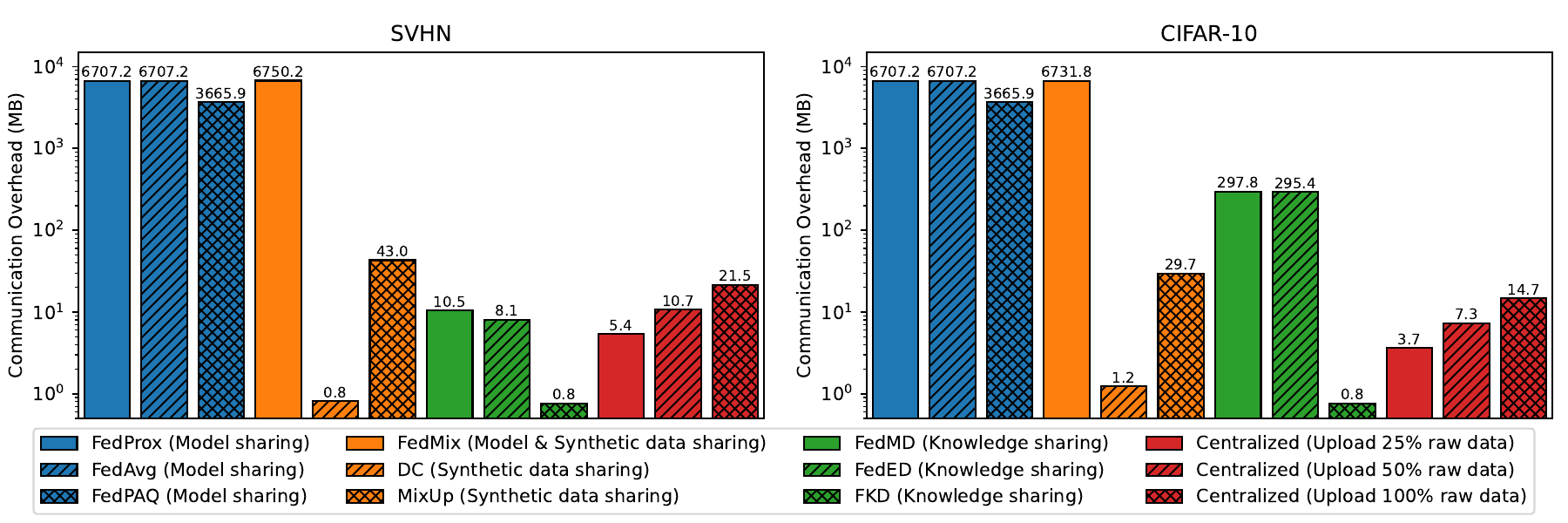}
    \caption{Communication overhead during the  training process.}
    \label{fig:communication_overhead}
\end{figure*}

\subsection{Communication Overhead}

In this subsection, we compare the communication overhead of different sharing methods, and Fig. \ref{fig:communication_overhead} shows the results.
Although model sharing methods achieve satisfactory test accuracy, as verified in Section \ref{sec:subsec:exp_privacy_utility}, they lead to higher communication overhead than data sharing and knowledge sharing methods.
This is mainly due to the large volume of local model parameters.
Particularly, FedPAQ offers a flexible balance between communication overhead and model performance. 
By quantizing the model updates, FedPAQ manages to reduce communication costs, but it causes a slight decrease in model performance.
For data sharing methods, the communication overhead is directly proportional to the size of the data being shared. Similarly, for data-based knowledge sharing methods like FedMD and FedED, the communication costs primarily depend on the quantity of proxy data used for transferring knowledge.
In the experiments, we utilize the MNIST and Tiny ImageNet datasets to transfer knowledge for the SVHN and CIFAR-10 classification tasks, respectively.
Since the gray-scale digits from MNIST are of smaller sizes than the RGB images from Tiny ImageNet, the communication overhead of both FedMD and FedED for the SVHN classification task is lower than that for the CIFAR-10 classification task.
Moreover, FKD exhibits the least communication overhead among the methods evaluated, because it does not rely on extra data for knowledge transfer.

\begin{figure}[t]
    \centering
    \includegraphics[width =0.6 \linewidth]{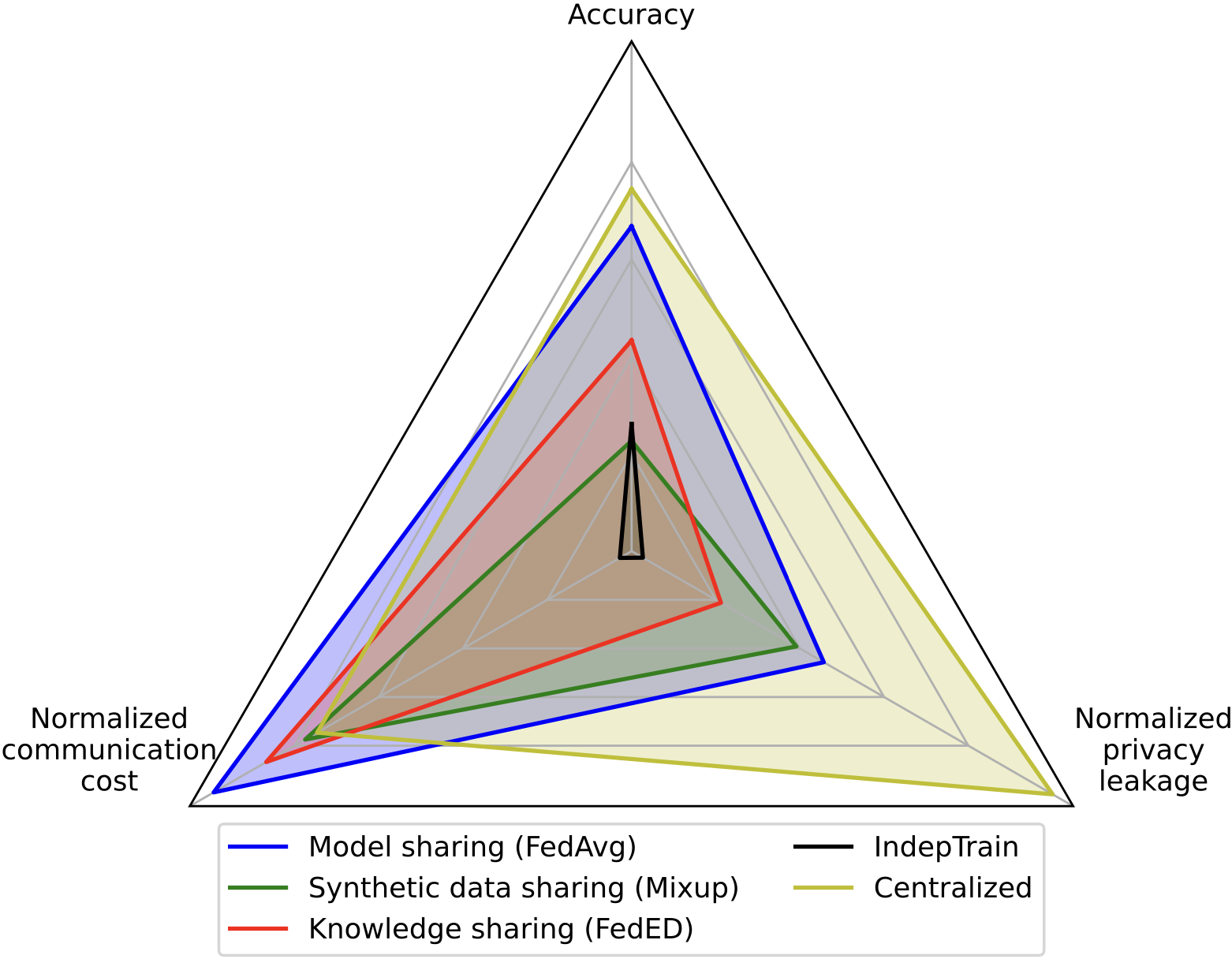}
    \caption{The accuracy, communication cost, and privacy leakage of different FL methods in the CIFAR-10 classification task with $\alpha = 0.1$. The communication costs are normalized by the logarithmic scale. 
    The normalized privacy leakage is a weighted sum of the membership inference accuracy and the FID value.}
    \label{fig:radar_figure}
\end{figure}

\subsection{Comparisons among Different Types of Sharing Methods}

Sharing different types of information influences the model utility, privacy leakage, and communication efficiency.
Fig. \ref{fig:radar_figure} presents a comparative analysis of five representative methods.
The centralized training scheme achieves the highest accuracy, but it leads to severe privacy problems.
In contrast, the model sharing method achieves comparable accuracy to the centralized training scheme with relatively low privacy leakage, but it incurs high communication overhead.
The knowledge sharing approach effectively balances accuracy, privacy preservation, and communication efficiency. 
Although this method exhibits slightly lower accuracy compared to the model sharing approach, it benefits from reduced communication overhead and a lower risk of privacy breaches. 
On the other hand, the state-of-the-art synthetic data sharing method achieves the worst performance. 
It poses a high risk of privacy leakage and considerable communication overhead, while its accuracy is lower than that of IndepTrain.
Given these findings, future research endeavors should prioritize the development of synthetic data generation techniques to improve the utility and minimize privacy concerns.

\subsection{Effectiveness of Defense Methods}

Finally, we evaluate the effectiveness of the defense methods in enhancing privacy protection against potential attacks.
The techniques evaluated in the experiments include the DP mechanism and data perturbation.

\subsubsection{DP Mechanism}

We employ the differentially-private stochastic gradient descent (DP-SGD) algorithm \cite{abadi2016deep} and the Opacus package \cite{yousefpour2021opacus} to optimize local models in four FL methods: FedAvg, FedPAQ, FedMD, and FedED.
The maximum norm of the gradient is set to 1, and the noise multiplier is set to 0.01. 
As shown in Table \ref{table:exp_dp}, the DP mechanism effectively protects the model against the membership inference attack but at the cost of reduced test accuracy.
This is because adding DP noise to the models makes it challenging for both clients and the attacker to distinguish between signal and random fluctuations.
However, the DP mechanism provides limited defense against the model inversion attack.
The empirical results presented in Table \ref{table:exp_dp} are consistent with the findings in \cite{zhang2020secret}, where the FID value may not increase after introducing the DP noise.
The authors of \cite{zhang2020secret} attributed these results to the fact that DP is mainly designed to protect the presence of a single data sample and not explicitly to protect attribute privacy.

\subsubsection{Data Perturbation}
To present effective defense methods against privacy attacks while avoiding negative impacts on the utility, recent studies have adopted data perturbation to defend against the membership inference attacks by mitigating the overfitting effect of local training \cite{sablayrolles2019white}.
In this part, we choose three representative methods for evaluation \cite{GA, RandomCrop}, including Gaussian augmentation, RandomCrop, and RandomFlip. 
Specifically, Gaussian augmentation incorporates the Gaussian noise into the input data for randomized smoothing.
RandomCrop crops the images at a random location, while RandomFlip randomly flips the input image horizontally or vertically.
Table \ref{table:data_aug} presents the test accuracy and membership inference accuracy of FL methods with or without data perturbation. 
It is observed that all data perturbation methods provide better privacy guarantees against membership inference attacks. 
Additionally, while RandomCrop and RandomFlip have the potential to improve the model utility, Gaussian augmentation results in a decrease in test accuracy, since directly adding Gaussian noise into the training samples degrades the image quality.
In conclusion, this set of experiments suggests that while data perturbation techniques are effective in mitigating privacy risks, they have varying effects on training performance.

\begin{table*}[]
\centering
\footnotesize
\caption{Test accuracy, membership inference (MEM) accuracy, and FID value of representative methods with and without DP protection.
The symbols $\Uparrow$ and $\Downarrow$ respectively denote higher and lower values are preferable.
The symbols \textcolor{green}{$\uparrow$} and \textcolor{red}{$\downarrow$} represent the increase or decrease in values compared to cases without DP protection.
}
\begin{tabular}{c|cc|cc|cc|cc}
\hline
\multirow{2}{*}{} & \multicolumn{2}{c|}{FedAvg} & \multicolumn{2}{c|}{FedPAQ} & \multicolumn{2}{c|}{FedMD} & \multicolumn{2}{c}{FedED} \\
\cline{2-9}
                            & \multicolumn{1}{c}{w/o DP}        & w/ DP       & \multicolumn{1}{c}{w/o DP}        & w/ DP       & \multicolumn{1}{c}{w/o DP}        & w/ DP      & \multicolumn{1}{c}{w/o DP}        & w/ DP      \\ \hline
Test $\Uparrow$                   &       66.81       &      61.19  \textcolor{red}{$\downarrow$}      &       58.82       &       56.66 \textcolor{red}{$\downarrow$}      &       47.51       &     38.91 \textcolor{red}{$\downarrow$}       &      43.35        &      35.93 \textcolor{red}{$\downarrow$}      \\
MEM $\Downarrow$                         &      65.50        &      52.11 \textcolor{red}{$\downarrow$}       &       57.13       &      51.30 \textcolor{red}{$\downarrow$}       &        51.75      &      50.78 \textcolor{red}{$\downarrow$}       &  50.80            &      50.68 \textcolor{red}{$\downarrow$}      \\
FID $\Uparrow$                         &     238.38         &       249.20 \textcolor{green}{$\uparrow$}      &       278.74      &      272.40 \textcolor{red}{$\downarrow$}      &       362.38       &      344.00 \textcolor{red}{$\downarrow$}     &      354.99        &     379.70 \textcolor{green}{$\uparrow$}   \\ \hline
\end{tabular}
\label{table:exp_dp}
\end{table*}

\begin{table*}[]
\centering
\caption{
Test accuracy and membership inference (MEM) accuracy of representative methods with data perturbation.
The symbols $\Uparrow$ and $\Downarrow$ respectively denote higher and lower values are preferable.
The symbols \textcolor{green}{$\uparrow$} and \textcolor{red}{$\downarrow$} represent the increase or decrease in values compared to the baseline.
}
\resizebox{0.98\textwidth}{!}{
\begin{tabular}{c|cccc|cccc}
\hline
\multicolumn{1}{c|}{\multirow{2}{*}{Accuracy}} & \multicolumn{4}{c|}{FedAvg}                                                                                                              & \multicolumn{4}{c}{FedPAQ}                                                                                                              \\ \cline{2-9} 
\multicolumn{1}{c|}{}                            & \multicolumn{1}{c}{Baseline} & \multicolumn{1}{c}{Gaussian Augmentation} & \multicolumn{1}{l}{RandomCrop} & \multicolumn{1}{l|}{RandomFlip} & \multicolumn{1}{c}{Baseline} & \multicolumn{1}{c}{Gaussian Augmentation} & \multicolumn{1}{l}{RandomCrop} & \multicolumn{1}{l}{RandomFlip} \\ \hline
\multicolumn{1}{c|}{Test $\Uparrow$}                    &       66.81                      &                       59.72 \textcolor{red}{$\downarrow$}                &               73.15 \textcolor{green}{$\uparrow$}                  &                         70.45 \textcolor{green}{$\uparrow$}        &               58.82              &             47.31 \textcolor{red}{$\downarrow$}                          &             61.25 \textcolor{green}{$\uparrow$}                     & \multicolumn{1}{c}{61.45 \textcolor{green}{$\uparrow$} }            \\ 
     \multicolumn{1}{c|}{MEM $\Downarrow$}                                          &        65.50                     &         63.72 \textcolor{red}{$\downarrow$}                               &           53.45 \textcolor{red}{$\downarrow$}                     &              62.06 \textcolor{red}{$\downarrow$}                  &            57.13                 &                    52.02 \textcolor{red}{$\downarrow$}                   &               51.27 \textcolor{red}{$\downarrow$}                 & \multicolumn{1}{c}{52.54 \textcolor{red}{$\downarrow$}}            \\ \hline \hline
\multirow{2}{*}{Accuracy}                       & \multicolumn{4}{c|}{FedMD}                                                                                                                & \multicolumn{4}{c}{FedED}                                                                                                                \\ \cline{2-9}
                                                  & \multicolumn{1}{c}{Baseline}                     & \multicolumn{1}{c}{Gaussian Augmentation}                    &  \multicolumn{1}{l}{RandomCrop}                      & RandomFlip                      &     \multicolumn{1}{c}{Baseline}                  & \multicolumn{1}{c}{Gaussian Augmentation}  &   \multicolumn{1}{c}{RandomCrop}                    & \multicolumn{1}{c}{RandomFlip}                       \\ \hline
 Test $\Uparrow$                                         & \multicolumn{1}{c}{47.51}        & \multicolumn{1}{c}{31.76 \textcolor{red}{$\downarrow$}}                   &              48.66 \textcolor{green}{$\uparrow$}                    &     48.29 \textcolor{green}{$\uparrow$}                             & \multicolumn{1}{c}{43.35}        &          36.90 \textcolor{red}{$\downarrow$}                              &               42.98 \textcolor{red}{$\downarrow$}                  &             43.79 \textcolor{green}{$\uparrow$}                    \\
MEM $\Downarrow$                                              & \multicolumn{1}{c}{51.75}        & \multicolumn{1}{c}{50.61 \textcolor{red}{$\downarrow$}}                   &                50.49 \textcolor{red}{$\downarrow$}                 &      50.68 \textcolor{red}{$\downarrow$}                          & \multicolumn{1}{c}{50.80}        &         50.38 \textcolor{red}{$\downarrow$}                              &              50.14 \textcolor{red}{$\downarrow$}                  &          50.31 \textcolor{red}{$\downarrow$} \\ \hline                     
\end{tabular}}
\label{table:data_aug}
\end{table*}

\section{Future Research Directions}

\label{sec:discussion}

After conducting a comprehensive review of what to share in FL and analyzing extensive experimental results, we have made several important observations and valuable findings.
However, despite the rapid development of FL in recent years, this research field is still facing challenges for further research.
Some open problems and promising directions are listed below:

\begin{itemize}

\item \textbf{How to effectively aggregate local information?}
This survey provided an overview of what to share in FL.
A complementary aspect that is equally important is the aggregation of shared information.
Model averaging is the most common approach, which has demonstrated the ability to produce a generalized global model.
However, such an approach faces problems in heterogeneous FL.
When clients have various hardware capabilities and network connectivity, straightforward model aggregation leads to low training efficiency and suboptimal convergence, since it requires local models to follow the same architecture.
Recent studies \cite{diao2020heterofl, hong2022efficient} attempted to address this issue by enabling resource-constrained clients to train only a subset of the global model. 
Nevertheless, the effectiveness of these methods may be limited by the statistical heterogeneity in local datasets.
As discussed in Section \ref{sec:sharing_method}, generating synthetic data and knowledge is naturally model-agnostic, and sharing them can mitigate the non-IID issue.
Therefore, investigating the hybrid aggregation mechanisms that jointly aggregate different types of information, such as model updates, synthetic data, and intermediate features, is a promising research direction for future work.

\item \textbf{How to improve the defense against privacy attacks?} As empirically evaluated in Section \ref{sec:exp}, more effective defense methods are often accompanied by a performance drop. 
For instance, DP-based methods typically add a substantial amount of noise to ensure data privacy, while significantly impacting the resulting model accuracy \cite{rodriguez2023survey}.
Besides, defense methods may differ in terms of effectiveness under different privacy attacks.
DP mechanisms successfully defend against membership inference attacks, but they often fail to counter model inversion attacks.
A promising solution is to develop hybrid privacy-preserving mechanisms that combine the strengths of various defense methods to mitigate their weaknesses.
In addition, explicitly quantifying the privacy guarantees provided by defense methods is essential to better understand their limitations and analyze the tradeoffs between model utility and privacy leakage.

\item \textbf{How to achieve communication-efficient FL?}
The convergence of federated training relies on the frequent information exchange between clients and the server, which undoubtedly results in considerable communication costs.
This problem is exacerbated when local resources are highly heterogeneous.
Different clients may have varying communication resources and power supplies, resulting in some clients taking considerably more time to complete the training round.
Besides compression methods reviewed in Section \ref{sec:sharing_method}, developing selection strategies is promising to improve efficiency.
First, the server dynamically selects a subset of clients that are likely to complete their updates on time in each training round, taking into account client availability, historical response times, and network connectivity. 
Second, the clients should prioritize transmitting the most impactful information to the server, such as significant parameter changes or highly informative synthetic samples.
As FL systems grow in scale, the development of such selection strategies becomes increasingly crucial.

\end{itemize}

\section{Conclusions}

As an innovative solution to address the privacy-preserving requirements of cutting-edge AI techniques, the emerging FL  paradigm still faces multiple challenges, including data heterogeneity, vulnerability to privacy attacks, and high communication overhead.
While recent survey articles have summarized the state-of-the-art advances in FL, most of them primarily focused on sharing model parameters during the training process and overlooked other sharing methods.
To fill this gap, this survey presented a new taxonomy that classifies FL methods into three categories based on the type of shared information, including model sharing, synthetic data sharing, and knowledge sharing.
We provided a comprehensive review of different sharing methods by analyzing their model utility, privacy leakage, and communication efficiency.
Moreover, we conducted extensive experiments to compare the performance of these methods, which yielded valuable insights and lessons.
Finally, we provided promising directions and open problems for further research.
This survey aims to serve as a comprehensive guide for researchers, facilitating their understanding of the latest developments and inspiring future studies.

\label{sec:conclusion}

\begin{acks}
This work was supported by the Hong Kong Research Grants Council under the Areas of Excellence scheme grant AoE/E-601/22-R and NSFC/RGC Collaborative Research Scheme grant CRS\_HKUST603/22.
\end{acks}

\bibliographystyle{ACM-Reference-Format}
\bibliography{ACMbibliography}


\end{document}